%% file: main.tex
\documentclass[10pt,journal,compsoc]{IEEEtran}
\ifCLASSOPTIONcompsoc
  \usepackage[nocompress]{cite}
\else
  \usepackage{cite}
\fi

\usepackage{wrapfig}
\usepackage{amsfonts}       
\usepackage{nicefrac}       
\usepackage{graphicx}
\usepackage{paralist}
\usepackage{nicefrac}       
\usepackage{microtype}      
\usepackage{xcolor}         
\usepackage{bm}
\usepackage{makecell}
\usepackage{bbm}
\usepackage{xspace}
\usepackage{float}
\usepackage{color}
\usepackage{colortbl}
\usepackage{dsfont}
\usepackage{amsmath,amssymb}
\usepackage{enumitem}   

\usepackage{subcaption}
\usepackage{epsfig}

\usepackage{wrapfig}
\usepackage{amsfonts}       
\usepackage{nicefrac}       
\usepackage{caption}
\usepackage{microtype}      
\usepackage{bm}
\usepackage{bbm}
\usepackage{xspace}
\usepackage{color}
\usepackage{colortbl}
\usepackage{dsfont}
\usepackage{multirow}
\usepackage{newunicodechar}
\newunicodechar{，}{,}

\usepackage{amsmath,amssymb} %
\include{math_commands}

\definecolor{airforceblue}{rgb}{0.36, 0.54, 0.66}

\definecolor{Gray}{gray}{0.9}

\usepackage[edges]{forest}
\definecolor{lightcoral}{rgb}{0.94, 0.5, 0.5}
\definecolor{lightgreen}{rgb}{0.56, 0.93, 0.56}

\definecolor{brightlavender}{rgb}{0.75, 0.58, 0.89}
\definecolor{capri}{rgb}{0.0, 0.75, 1.0}
\definecolor{carminepink}{rgb}{0.92, 0.3, 0.26}
\definecolor{celadon}{rgb}{0.67, 0.88, 0.69}
\definecolor{darkpastelgreen}{rgb}{0.01, 0.75, 0.24}

\definecolor{pastelblue}{rgb}{0.68, 0.78, 0.81}
\definecolor{mintgreen}{rgb}{0.6, 0.98, 0.6}
\definecolor{lavender}{rgb}{0.71, 0.49, 0.86}
\definecolor{peach}{rgb}{1.0, 0.9, 0.71}
\definecolor{coral}{rgb}{1.0, 0.5, 0.31}
\definecolor{mauve}{rgb}{0.88, 0.69, 1.0}
\definecolor{lemonyellow}{rgb}{1.0, 0.96, 0.4}

\definecolor{hidden-draw}{RGB}{205, 44, 36}
\definecolor{hidden-blue}{RGB}{194,232,247}
\definecolor{hidden-orange}{RGB}{243,202,120}
\definecolor{hidden-yellow}{RGB}{242,244,193}
\definecolor{tree-level-1}{RGB}{245,20,85}
\definecolor{tree-level-2}{RGB}{246,86,118}
\definecolor{tree-level-3}{RGB}{248,177,193}
\definecolor{tree-leaf}{RGB}{176,230,198}

\definecolor{Self}{RGB}{255,0,128}
\definecolor{Ensemble}{RGB}{0,127,255}
\definecolor{Iterative}{RGB}{153,51,255}

\definecolor{exemplar1}{RGB}{136,98,148}
\definecolor{exemplar2}{RGB}{148,210,242}
\definecolor{knowledge1}{RGB}{249,219,152}
\definecolor{knowledge2}{RGB}{255,245,220}

\definecolor{lighttealblue}{RGB}{41, 157, 143}
\definecolor{lightplum}{RGB}{233, 196, 106}
\definecolor{harvestgold}{RGB}{216, 118, 89}

\usepackage{nicematrix}
\usepackage{makecell}
\usepackage{xspace,mfirstuc,tabulary}
\usepackage{booktabs}
\usepackage{graphicx}
\usepackage{bbding}
\usepackage{multirow}

\usepackage{tabularx}   
\usepackage{ragged2e}   

\usepackage{pifont}
\newrobustcmd{\B}{\bfseries}
\newcommand*\colourcheck[1]{%
  \expandafter\newcommand\csname #1check\endcsname{\textcolor{#1}{\ding{52}}}%
}
\newcommand*\colourcross[1]{%
  \expandafter\newcommand\csname #1cross\endcsname{\textcolor{#1}{\ding{55}}}%
}
\colourcheck{green}
\colourcross{red}
\newcommand{\eg}{\emph{e.g.}}
\newcommand{\ie}{\emph{i.e.}}

\renewcommand{\arraystretch}{1.2}  

\definecolor{mygray}{gray}{.92}

\renewcommand{\texttt}[1]{ $ {{\tt #1} } $}

\newcommand{\Mset}{\mathcal{M}_{\mathrm{set}}}     
\newcommand{\Msamp}{\mathcal{M}_{\mathrm{samp}}}   
\newcommand{\Mstate}{\mathcal{M}_{\mathrm{state}}} 

\usepackage[ruled]{algorithm2e}         
\usepackage{array}
\newcolumntype{I}{!{\vrule width 3pt}}
\newlength\savedwidth

\newlength\savewidth

\renewcommand{\texttt}[1]{ $ {{\tt #1} } $}  

\usepackage{xcolor}
\newcommand{\secrefblue}[1]{\textcolor{blue}{\S\ref{#1}}}

\def\eg{\emph{e.g., }}

\def\ie{\emph{i.e., }}

\usepackage[breaklinks=true,bookmarks=false]{hyperref}
\hypersetup{
colorlinks=true,
linkcolor=black
}
\definecolor{burlywood}{RGB}{222,184,135}
\definecolor{wheat}{RGB}{245,222,179}
\definecolor{lightsteelblue}{RGB}{176,196,222}

\begin{document}
%
\title{Reinforcement Learning for Large Model: A Survey}

\author{
{\small
Weijia Wu$^1$},
{\small
Chen Gao$^{1}$},
{\small
Joya Chen$^{1}$},
{\small
Kevin Qinghong Lin$^{1}$},
{\small
Qingwei Meng$^{2}$},
{\small
Yiming Zhang$^{3}$}\\,
{\small
Yuke Qiu$^{2}$},
{\small
Hong Zhou$^{2}$},
{\small
Mike Zheng Shou$^1$$^\ddagger$\thanks{ 
Corresponding author 
($^\ddagger$)
: Mike Zheng Shou (mikeshou@nus.edu.sg).}}\\
$ \qquad $\\
{\normalsize
$^1$National University of Singapore$ \qquad $ $^2$Zhejiang University $ \qquad $ $^{3}$The Chinese University of Hong Kong }\\
}

\IEEEtitleabstractindextext{%
\input{0_abstract}

}


\maketitle
\tableofcontents
\newpage 
\vspace*{2\baselineskip}

\input{1_introduction}

\input{2_preliminary}

\input{3_main_content}

\input{4_benchmark_and_metric}
\input{5_future}

\input{6_conclusion}

\IEEEdisplaynontitleabstractindextext

%
\IEEEpeerreviewmaketitle







\bibliographystyle{IEEEtran}
\bibliography{egbib}

\end{document}

%% file: 0_abstract.tex
\begin{abstract}
Recent advances at the intersection of reinforcement learning (RL) and visual intelligence have enabled agents that not only perceive complex visual scenes but also reason, generate, and act within them.
This survey offers a critical and up-to-date synthesis of the field.
We first formalize visual RL problems and trace the evolution of policy-optimization strategies from RLHF to verifiable reward paradigms,
and from Proximal Policy Optimization to Group Relative Policy Optimization.
We then organize more than $250$ representative works into four thematic pillars: multi-modal large language models, visual generation, unified model frameworks, and vision-language-action models.
For each pillar we examine algorithmic design, reward engineering, benchmark progress, and we distill trends such as curriculum-driven training, preference-aligned diffusion, and unified reward modeling. 
Finally, we review evaluation protocols spanning set-level fidelity, sample-level preference, and state-level stability, and we identify open challenges that include sample efficiency, generalization, and safe deployment.
Our goal is to provide researchers and practitioners with a coherent map of the rapidly expanding landscape of visual RL and to highlight promising directions for future inquiry.
Resources are available at: \href{https://github.com/weijiawu/Awesome-Visual-Reinforcement-Learning}{\color{blue}{\tt Visual RL}}.

\end{abstract}

%% file: 1_introduction.tex

\section{Introduction}\label{sec:introduction}

Reinforcement Learning (RL) has achieved remarkable successes in the field of Large Language Models (LLMs)~\cite{jaech2024openai,rafailov2023direct}, most notably through paradigms like Reinforcement Learning from Human Feedback (RLHF)~\cite{ouyang2022training} and innovative frameworks such as DeepSeek-R1~\cite{guo2025deepseek}.
These methodologies have significantly enhanced the capabilities of LLMs, aligning generated outputs closely with human preferences and enabling nuanced, complex reasoning and interaction capabilities previously unattainable through supervised learning alone.

\begin{figure*}[t]
    \centering
	\includegraphics[width=0.98\linewidth]{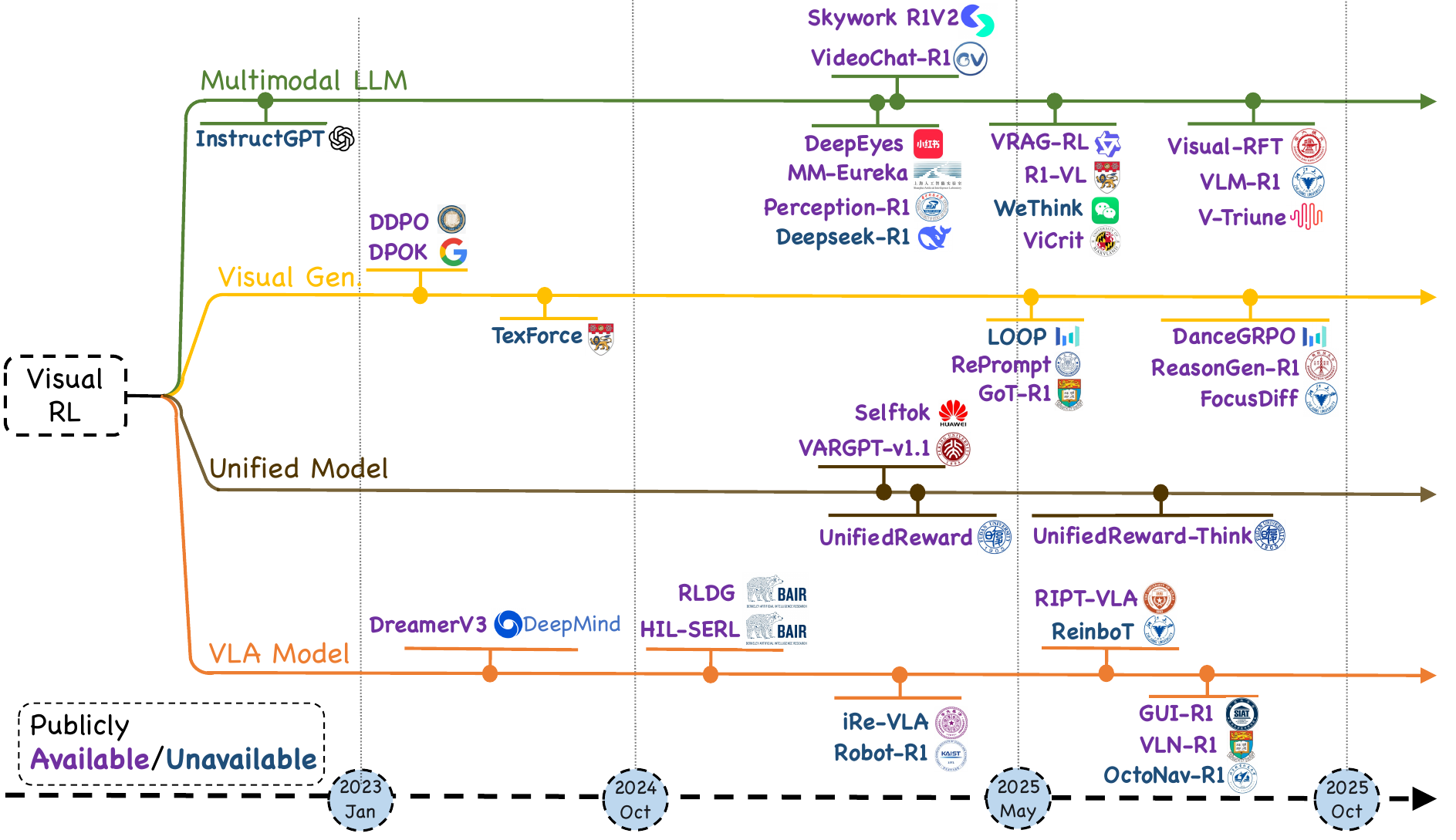}
	\caption{\textbf{Timeline of Representative Visual Reinforcement Learning Models.} 
     The figure presents a chronological overview of key Visual RL models from 2023 to 2025, organized into four tracks: Multimodal LLM, Visual Generation, Unified Models, and VLA Models. 
    }

    \vspace{-0.1cm}
\label{Timeline}
\end{figure*}

In recent years, inspired by these remarkable achievements from LLM, there has been an explosive interest in extending the RL methodologies that proved successful for LLMs to multimodal large models, including Vision-Language Models (VLM)~\cite{zhou2025reinforced,zheng2025deepeyes,zhang2025chain}, Vision-Language-Action models (VLA)~\cite{lu2025vla,tan2025interactive,luo2025gui,yuan2025enhancing}, diffusion-based visual generation models~\cite{fan2023reinforcement,black2023ddpo,zhou2025dreamdpo}, and unified multimodal frameworks~\cite{mao2025unirl,wang2024emu3,wang2025discrete}, as shown in Figure~\ref{Timeline}.
Multimodal models such as Gemini 2.5~\cite{google_gemini2.5_2025} have leveraged RL to align visual-textual reasoning processes and produce outputs with higher semantic coherence and alignment with human judgments. 
Concurrently, VLA models integrating vision and language with action-oriented outputs have adopted RL to optimize complex sequential decision-making processes in interactive environments, significantly improving task-specific performance in GUI automation~\cite{yuan2025enhancing,shi2025mobilegui}, robotic manipulation~\cite{lu2025vla}, and embodied navigation~\cite{kim2025rapid}.
The rapid advancement of diffusion-based generative models has further spurred this RL-driven innovation wave.
Works like ImageReward~\cite{xu2023imagereward} have introduced reinforcement learning to enhance the semantic alignment and visual quality of generative outputs, refining diffusion-based generation through iterative feedback mechanisms derived from human preferences or automated reward critics. 
Moreover, unified models that blend multiple tasks, \ie understanding, and generation, into single architectures~\cite{mao2025unirl,jiang2025co} have increasingly relied on RL-driven fine-tuning, achieving generalization and task transfer previously considered challenging.
Despite the substantial progress in integrating reinforcement learning with multimodal large language models, several core challenges remain.
These include stabilizing policy optimization under complex reward signals, managing high-dimensional and diverse visual inputs, and designing scalable reward functions that support long-horizon decision-making.
Addressing these challenges necessitates methodological innovations in both algorithmic design and evaluation protocols.

In this survey, we present a comprehensive synthesis of recent advances in visual reinforcement learning within the context of multimodal large models, with a focus on the surge of research activity since 2024.
We begin by revisiting foundational RL successes in language models, such as RLHF~\cite{ouyang2022training} and DeepSeek-R1~\cite{guo2025deepseek}, which have laid the groundwork for multimodal adaptation.
Subsequently, we discuss how these strategies have evolved in the visual domain, categorizing over 200 representative works into four key domains: (i) multimodal large language models, (ii) visual generation, (iii) unified RL frameworks, and (iv) vision-language-action agents, as shown in Figure~\ref{Timeline}.
Within each category, we analyze key developments in algorithmic formulations, reward modeling, and benchmarking methodologies.
Finally, we identify open challenges and future directions, highlighting the need for more efficient multimodal reasoning, robust long-horizon learning strategies for VLA tasks, and scalable, high-fidelity reward signals tailored to visual generation.
Through this comprehensive overview, we offer a structured overview of visual reinforcement learning to support future research and practical deployment in this rapidly evolving field.

Our key contributions are as follows:
\begin{itemize}
    \item We provide a systematic and up-to-date survey of over 200 visual reinforcement learning studies, encompassing MLLMs, visual generation, unified models, and vision-language-action agents.
    
    \item We analyze advances in policy optimization, reward modeling, and benchmarking across subfields, revealing key challenges and future directions: such as reward design in visual generation and the lack of intermediate supervision in reasoning and VLA tasks.
    
    \item We introduce a principled taxonomy of Visual RL methods based on metric granularity and reward supervision, including three reward paradigms for image generation. 
    This framework clarifies the design trade-offs across domains and offers actionable insights for selecting and developing RL strategies.
\end{itemize}

%% file: 2_preliminary.tex
\section{Preliminary: Reinforcement Learning in LLM}
\label{sec:2_reinforcement_learning}

\begin{table*}[t]
    \centering
    
    \label{tab:ppo_symbols}
    \setlength{\tabcolsep}{1mm}
    \caption{\textbf{Glossary of Symbols for Visual Reinforcement Learning.} It consolidates the notation that recurs across Sections~\secrefblue{subsec:notation}–\secrefblue{subsec:PolicyOptimization}, with the rightmost column pointing to each appearance of the symbol.}
    \input{symbol}
\end{table*}

This section lays the foundation for the RL of multi-modal models.
We first formalize the notation (\secrefblue{subsec:notation}), casting text and image generation as a Markov Decision Process.
Next, we examine three alignment paradigms (\secrefblue{subsec:Alignment}): \emph{RL from Human Feedback} (RLHF), \emph{Group-Relative Policy Optimization} (GRPO), and \emph{Reinforcement Learning with Verifiable Rewards} (RLVR), each aligning policies via human preferences or deterministic checks.
Finally, \secrefblue{subsec:PolicyOptimization} reviews the core policy gradient methods (PPO, GRPO) and their adaptations to visual reasoning and generation.


\subsection{Notation and Problem Formulation}
\label{subsec:notation}
We cast text- or image-generation as an episodic Markov decision process.  
We treat the user \emph{prompt} \(p\) as the \textbf{initial state}
$s_0=p$.
At timestep \(t\), the state is the prompt plus all previously generated
actions:
\begin{align}
  s_t \;=\; \bigl(p,\,a_1,\dots,a_{t-1}\bigr).
\end{align}
A \emph{continuation} is the full action sequence
$ \{a_1,\ldots,a_T\}$,
where each token \(a_t \in \mathcal{A}\) is sampled autoregressively from the policy:
\begin{align}
  \pi_\theta\!\bigl(a_t \mid s_t\bigr)
  \;=\;
  \pi_\theta\!\bigl(a_t \mid p,a_1,\dots,a_{t-1}\bigr).
\end{align}
In words, the prompt anchors the state sequence, and each new action
is chosen in the context of that prompt and the tokens already produced.

A fixed reference
model (\textit{e.g.,} the supervised fine-tuned checkpoint) is denoted
$\pi_{\text{ref}}$.  Human preferences are distilled into a scalar reward model
$R_\phi(p,y)$, replacing the reward from the unknown environment.
We write
$\rho_t(\theta)=\pi_\theta(a_t\!\mid\!s_t)\big/\pi_{\theta_{\text{old}}}(a_t\!\mid\!s_t)$  
for the importance ratio between new and behavior policies.
$\hat A_t$ is used to denote the advantage estimate.  
PPO reduces the variance of $\hat A_t$ with a learned critic
$\hat V_\psi$, whereas GRPO replaces the critic by a
\emph{group-relative} baseline computed from a set
$O=\{a_i\}_{i=1}^{G}$ of continuations that share the same prompt.
All two algorithms add a KL regulariser
$\text{KL}\!\bigl(\pi_\theta(\cdot\!\mid\!p)\Vert\pi_{\text{ref}}(\cdot\!\mid\!p)\bigr)$
weighted by~$\beta$ to keep the updated policy close to the reference.
Unless stated otherwise, expectations $\mathbb E[\cdot]$ are over
prompts $p\!\sim\!\mathcal D$ and continuations drawn from the specified policy.

\begin{figure*}[t]
    \centering
	\includegraphics[width=0.98\linewidth]{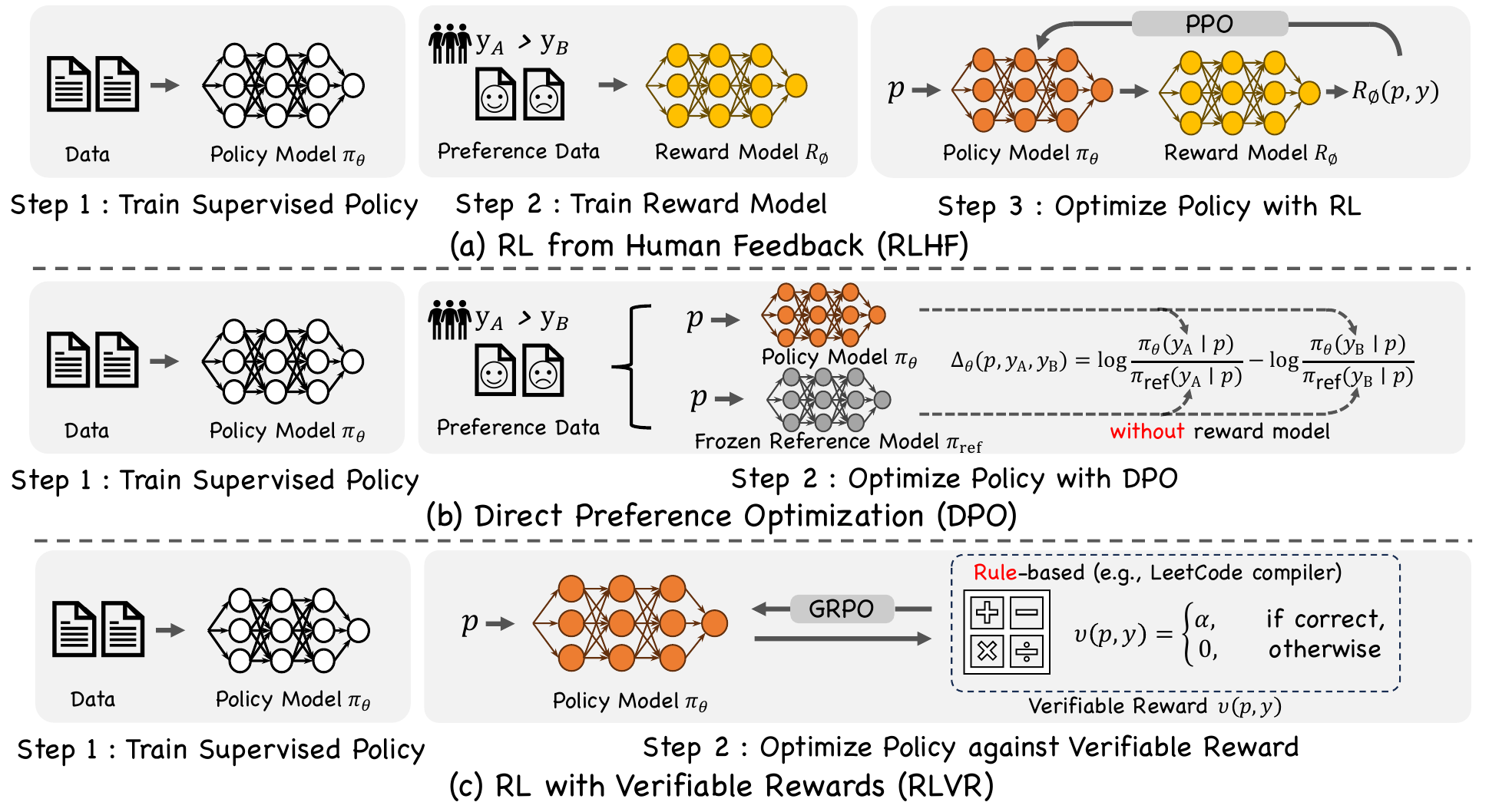}
	\caption{\textbf{Three Alignment Paradigms for Reinforcement Learning.}
     (a) RLHF learns a reward model from human preference data and optimizes the policy via PPO.  
    (b) DPO removes the reward model and directly optimizes a contrastive objective against a frozen reference.  
    (c) RLVR replaces subjective preferences with deterministic verifiable signals and trains the policy using GRPO.
    }
\label{AlignmentParadigms}
\end{figure*}

\subsection{Alignment Paradigms}
\label{subsec:Alignment}
\subsubsection{RL from Human Feedback}
\label{subsec:rlhf}

RLHF~\cite{ouyang2022training} extends the underlying MDP with \emph{pairwise preference data}
curated from human annotators.  
Each preference example is a triple
$(p,\,y_{\mathrm{A}},\,y_{\mathrm{B}})$,
where $p$ is the prompt (or state sequence) and
$(y_{\mathrm{A}},y_{\mathrm{B}})$ are two candidate continuations
(trajectories, images, \emph{etc.});
the label $y\!\in\!\{0,1\}$ records which continuation is preferred, as shown in Figure~\ref{AlignmentParadigms}.

\textbf{Reward-model learning.}
A scalar reward model $R_\phi$ is trained to reproduce the pairwise
ordering via a Bradley–Terry likelihood:
\begin{align}
\mathcal{L}_{\text{RM}}
  &= -\!\!\sum_{(p,y_{\mathrm{A}},y_{\mathrm{B}})}
     \Bigl[
        y\,\log\sigma\!\bigl(
          R_\phi(p,y_{\mathrm{A}})-R_\phi(p,y_{\mathrm{B}})
        \bigr)
        \nonumber\\[-4pt]
  &\quad +(1-y)\,\log\sigma\!\bigl(
          R_\phi(p,y_{\mathrm{B}})-R_\phi(p,y_{\mathrm{A}})
        \bigr)
     \Bigr],
\end{align}
where $\sigma(\cdot)$ is the logistic function.
After convergence, $R_\phi$ provides a \emph{dense, differentiable}
proxy for human preference.

\textbf{Policy optimization.}
The policy $\pi_\theta$ is finally fine-tuned by maximizing  
(i) the learned reward,  
(ii) a KL penalty that keeps the policy close to the supervised-fine-tuned baseline $\pi_{\text{SFT}}$, and  
(iii) an \emph{optional} log-likelihood regulariser on the original pre-training distribution, as introduced in InstructGPT~\cite{ouyang2022training}\footnote{The coefficients
$\beta$ and $\gamma$ respectively control the strength of the KL penalty and the
pre-training log-likelihood term.  Setting $\gamma\!=\!0$ recovers the standard PPO objective.}:

\begin{align}
\max_{\theta}\;&
\underbrace{\mathbb{E}_{(p,y)\sim\pi_\theta}\!
  \bigl[R_\phi(p,y)\bigr]}_{\text{reward}}
-\beta\,
\underbrace{\mathbb{E}_{p}\!
  \bigl[\mathrm{KL}\!\bigl(
      \pi_\theta(\cdot\!\mid\!p)\;\|\;\pi_{\text{SFT}}(\cdot\!\mid\!p)
  \bigr)\bigr]}_{\text{SFT\,anchoring}}
\nonumber\\[-3pt]
&\quad
+\;\gamma\,
\underbrace{\mathbb{E}_{x\sim D_{\text{pretrain}}}
  \bigl[\log\pi_\theta(x)\bigr]}_{\text{pre-training log-likelihood}}\;.
\label{eq:ppo_kl_ptx}
\end{align}

In practice, the first two terms are optimised with KL-regularised PPO over
minibatches of sampled continuations, while the third term adds the
pre-training gradients (“PPO-ptx” in \cite{ouyang2022training}) to mitigate
performance regressions on the original corpus.

\textbf{Three-stage recipe.}
Most modern RLHF pipelines follow the three-stage recipe, as shown in Figure~\ref{AlignmentParadigms} (a).
Step 1: Collect demonstration data, and train a supervised policy;
Step 2: Collect comparison data, and train a reward model;
Step 3: Optimize a policy $\pi_\theta$ against the reward model using PPO.
The paradigm was \emph{pioneered} by Christiano et al.,~\cite{christiano2017deep}, who trained Atari and robotic agents from pairwise human preferences.
Ouyang et al.,~\cite{ouyang2022training} later \emph{scaled} the recipe to large language models (InstructGPT) by coupling preference modeling with PPO.
For vision, reward models such as ImageReward~\cite{xu2023imagereward} and Human Preference Score (HPS)~\cite{wu2023human} supply dense aesthetic signals that guide text-to-image diffusion and related tasks.

\subsubsection{Direct Preference Optimization}
\label{subsec:dpo}

Direct Preference Optimisation (DPO)~\cite{rafailov2023direct}
takes exactly the \emph{same} pairwise-preference data as RLHF but
removes the intermediate reward-model and RL loop.
Instead, it derives a \emph{closed-form, supervised} objective that
implicitly enforces a KL constraint to a frozen reference
policy~$\pi_{\text{ref}}$, as shown in Figure~\ref{AlignmentParadigms} (b).

\textbf{Closed-form objective.}
For every prompt $p$ annotators rank two continuations
$\bigl(y_{\mathrm{A}},y_{\mathrm{B}}\bigr)$ and order them so that
$y_{\mathrm{A}}$ is the preferred continuation (“winner”) and
$y_{\mathrm{B}}$ the non-preferred one (“loser”).
Thus the dataset consists of triples
$(p,\,y_{\mathrm{A}},\,y_{\mathrm{B}})\!\sim\!\mathcal{D}$.
Let $\pi_{\text{ref}}$ be a frozen reference policy
(e.g.,\ the SFT checkpoint) and let $\beta>0$ be a temperature
hyper-parameter.
DPO minimizes:
\begin{align}
\mathcal{L}_{\text{DPO}}=
  -\,\mathbb{E}_{(p,y_{\mathrm{A}},y_{\mathrm{B}})\sim\mathcal{D}}
  \Bigl[
      \log\sigma\!\bigl(
          \beta\,\Delta_\theta(p,y_{\mathrm{A}},y_{\mathrm{B}})
      \bigr)
  \Bigr],
  \label{eq:dpo_loss}
\end{align}
where the log-odds gap is:
\begin{align}
\Delta_\theta(p,y_{\mathrm{A}},y_{\mathrm{B}})
  &= \log\frac{\pi_\theta(y_{\mathrm{A}}\mid p)}
                  {\pi_{\text{ref}}(y_{\mathrm{A}}\mid p)}
     \;-\;
     \log\frac{\pi_\theta(y_{\mathrm{B}}\mid p)}
                  {\pi_{\text{ref}}(y_{\mathrm{B}}\mid p)}
     \nonumber\\
  &= \bigl[\log\pi_\theta(y_{\mathrm{A}}\mid p)
           -\log\pi_\theta(y_{\mathrm{B}}\mid p)\bigr]
     \nonumber\\
  &\quad-\,
     \bigl[\log\pi_{\text{ref}}(y_{\mathrm{A}}\mid p)
           -\log\pi_{\text{ref}}(y_{\mathrm{B}}\mid p)\bigr].
\label{eq:dpo_gap}
\end{align}
The logistic function $\sigma(z)=1/(1+e^{-z})$
turns the gap into a binary-classification loss; 
training proceeds
with standard maximum-likelihood gradients, no reward model, value
network, or importance sampling is required.

\subsubsection{Reinforcement Learning with Verifiable Rewards}
\label{subsec1:rlvr}

Reinforcement Learning with Verifiable Rewards (RLVR) eliminates the
subjectivity and data-collection cost of RLHF by replacing pairwise human
preferences with \emph{deterministic, programmatically checkable} reward
signals \(v:\,(p,y)\!\mapsto\!\{0,1,\dots,K\}\).
Typical examples include \texttt{pass/fail} unit tests for code synthesis,
exact-match answers in mathematics, IoU/\textsc{dice} thresholds for
segmentation, or formal output-format validators (\textit{e.g.,} LeetCode compiler).
Because the reward is generated \emph{online} by execution or metric
evaluation, RLVR removes both  
(i) the reward-model training stage of RLHF and  
(ii) the contrastive surrogate loss of DPO, while still enabling substantial
policy improvements beyond supervised learning
\cite{guo2025deepseek,shao2024deepseekmath,lambert2024t},
as shown in Figure~\ref{AlignmentParadigms}\,(c).

\textbf{Verifiable reward.}
For a prompt (state)~\(p\) and a sampled continuation~\(y\), a verifier
returns:
\begin{equation}
  r(p,y)\;=\;v(p,y)\in\{0,1\},
\end{equation}
\textit{e.g.,} “\texttt{pass}” if the generated program solves all hidden tests.
The same idea applies to vision: a generated mask that attains
\(\mathrm{IoU}\!\ge\!0.9\) with ground truth, or an image whose CLIP similarity
exceeds a hard threshold can be awarded \(r\!=\!1\).
Current most RLVR systems (\eg~ DeepSeekMath, Deepseek-R1)
adopt GRPO~\cite{shao2024deepseekmath} (see Equ.~\ref{eq:grpo_obj}) and standard KL regularization to train the policy model.

RLVR therefore follows a concise two-stage pipeline.
Step 1: Supervised policy pre-training on demonstrations $\{(p,y)\}$, producing the initial $\pi_{\text{SFT}}$.
Step 2: RL fine-tuning with GRPO/PPO against the on-the-fly verifiable reward~$v(p,y)$, optionally mixing in a small percentage of teacher-forced SFT updates to stabilise training.

\begin{figure}[t]
    \centering
	\includegraphics[width=0.96\linewidth]{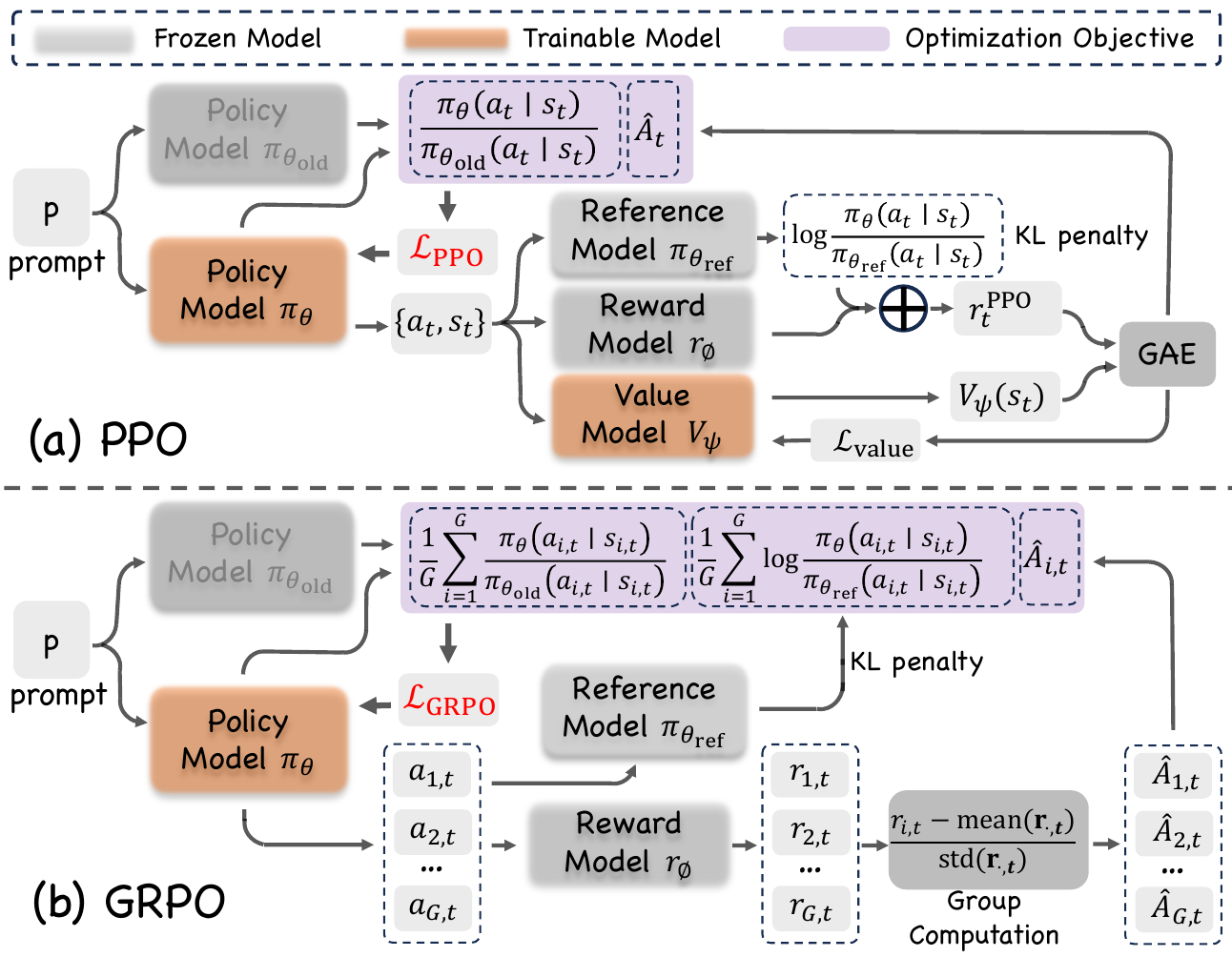}
	\caption{\textbf{Two Representative Policy Optimization Algorithms for LLM.}
     PPO (a) uses a learned value model $V_\psi$ for advantage estimation and injects the KL penalty at each token.
     GRPO (b) removes the value model, computes group-normalized advantages $\hat{A}_{i,t}$ across $G$ continuations, and applies an explicit prompt-level KL penalty.
    }
\label{PolicyOptimization}
\end{figure}

\subsection{Policy-Optimization Algorithms}
\label{subsec:PolicyOptimization}

\subsubsection{Proximal Policy Optimization}
\label{subsec:ppo}

Proximal Policy Optimization (PPO)~\cite{schulman2017proximal} is a first order
trust region method that updates the policy~$\pi_\theta$ while \emph{staying
close to the previous policy}~$\pi_{\theta_{\text{old}}}$ at every step,
as shown in Figure~\ref{PolicyOptimization} (a).
In text- or image-generation problems we treat a prompt~$p$ as the initial state
$s_0$ and the continuation~$\{a_1,\ldots,a_T\}$ as the trajectory.
Define the importance-sampling ratio $\rho_t(\theta)$ and immediate reward $r_\phi$:
\begin{itemize}
  \item \textbf{Importance–sampling ratio}
        \[
          \rho_t(\theta) \;=\;
          \frac{\pi_\theta(a_t \mid s_t)}
               {\pi_{\theta_{\text{old}}}(a_t \mid s_t)} ,
        \]
        which re-weights the gradient estimate from the behavior policy to the
        updated policy.
  \item \textbf{Immediate reward}
        \(
          r_\phi(s_t,a_t)
        \),
        provided by a \emph{frozen} reward model
        $r_\phi$ that has been
        pre-trained to approximate human preference.
  \item \textbf{Value baseline}
        \(
          V_\psi(s_t)
        \),
        produced by a \emph{learned} value network
        $V_\psi$ that regresses the expected
        discounted return from state~$s_t$.
\end{itemize}
With the KL-regularised reward between the policy model and reference model, $r_t^{\text{PPO}}$ can be defined:
\begin{align}
  r_t^{\text{PPO}}
  \;=\;
  r_\phi(s_t,a_t)
  \;-\;
  \beta\,
  \log\!
    \dfrac{\pi_\theta(a_t \mid s_t)}
          {\pi_{\text{ref}}(a_t \mid s_t)},
    \label{eq:ppoKL}
\end{align}
where the KL term(latter item) keeps the updated policy \(\pi_\theta\) from drifting too far
from the frozen reference \(\pi_{\text{ref}}\).
$\beta$ balances exploration (via KL proximity to the frozen
reference model $\pi_{\text{ref}}$) against exploitation of the reward model.
A larger \(\beta\) implies
stricter proximity and thus safer but potentially slower learning.
Then the generalized advantage estimator (GAE)~\cite{schulman2015high} produces $\hat{A}_t$:
\begin{align}
  \hat{A}_t
  \;=\;
  \operatorname{GAE}\!\bigl(r_t^{\text{PPO}},\,V_\psi\bigr),
  \label{eq:ppo_GAE}
\end{align}
where GAE computes advantage values by exponentially weighting multi-step reward estimates, providing a smooth trade-off between low-variance learning and high-variance Monte Carlo returns.

The surrogate objective maximized by PPO is then:
\begin{align}
  \mathcal{L}_{\text{PPO}}=
  \mathbb{E}_{t}\!\Bigl[
      \min\!\Bigl(
        \rho_t(\theta)\,\hat{A}_t,
        \operatorname{clip}\!\bigl(\rho_t(\theta),1-\epsilon,1+\epsilon\bigr)
        \,\hat{A}_t
      \Bigr)
  \Bigr],
  \label{eq:ppo}
\end{align}
where the hyper-parameter $\epsilon\!\in\!(0,1)$ controls the width of the trust region.
Accurate and low-variance \(\hat{A}_t\) estimates are therefore critical, they
direct each policy update and ultimately determine the stability and
sample efficiency of PPO.%

\subsubsection{Group Relative Policy Optimization}
\label{subsec:grpo}

Group Relative Policy Optimisation (GRPO)~\cite{shao2024deepseekmath}
extends PPO by discarding the learned value (critic) network and replacing
it with a \emph{group-relative} baseline computed from multiple outputs
sampled for the same prompt. 
This design markedly reduces memory
consumption while aligning the advantage estimator with the
comparison-based reward model, as shown in Figure~\ref{PolicyOptimization} (b).

\textbf{Group Relative Baseline.}
For each prompt~$p$,
we sample a group of $G$ full continuations 
$a_1,\dots,a_G \sim \pi_\theta(\cdot \mid p)$, 
where each continuation $a_{\cdot,t} = (a_{1,t}, \dots, a_{G,t})$ 
is a sequence of tokens indexed by timestep~$t$. 
The frozen reward model $r_\phi(p, a_{i,t})$ then assigns a scalar 
score to each token $a_{i,t}$ conditioned on the prompt.
These sequence level rewards are then normalized across the group 
to compute a group-relative advantage signal:
\begin{equation}
  \hat{A}_{i,t} = \tilde{r}_{i,t} = 
  \frac{r_{i,t} - \operatorname{mean}(\mathbf{r}_{\cdot,t})}
       {\operatorname{std}(\mathbf{r}_{\cdot,t})},
\end{equation}
where $\operatorname{mean}(\cdot)$ and $\operatorname{std}(\cdot)$ denote the mean and standard deviation functions used to compute the group relative advantage.
The same $\hat{A}_i$ is reused for every token
$a_{i,t}$ in the continuation, producing the clipped surrogate:
\begin{equation}
\label{eq:grpo_obj}
\begin{aligned}
\mathcal{L}_{\text{GRPO}} \;=\;&
\mathbb{E}_{p\sim\mathcal{D}}\Bigl[
      \tfrac{1}{G}\sum_{i=1}^{G}
      \tfrac{1}{|a_i|}\sum_{t=1}^{|a_i|}
      \min\!\bigl(
          \rho_{i,t}\,\hat{A}_{i,t}, \\[2pt]
&\operatorname{clip}\!\bigl(\rho_{i,t},1-\epsilon,1+\epsilon\bigr)\,
          \hat{A}_{i,t}
      \bigr)
    \Bigr] \\[4pt]
&-\beta\,\mathbb{E}_{p}\Bigl[
      D_{\mathrm{KL}}\Bigl(
        \pi_\theta(\cdot\mid p)\,\Big\|\,\pi_{\text{ref}}(\cdot\mid p)
      \Bigr)
    \Bigr],
\end{aligned}
\end{equation}
where
\(
  \rho_{i,t}
  =
  \pi_\theta(a_{i,t}\!\mid\!s_{i,t})
  \big/
  \pi_{\theta_{\text{old}}}(a_{i,t}\!\mid\!s_{i,t}).
\)
The explicit KL penalty $D_{\mathrm{KL}}(\cdot)$ keeps $\pi_\theta$ near the reference
$\pi_{\text{ref}}$, while the group-relative advantage $\hat{A}_i$ replaces
the value baseline $V_\psi$, roughly halving memory and compute yet retaining
a low-variance learning signal.

\textbf{Prompt-level KL estimator.}  
Instead of injecting a token-wise penalty into the reward (as PPO does with 
$\beta\log\frac{\pi_\theta}{\pi_{\text{ref}}}$), GRPO adds a \emph{separate} 
prompt-level regulariser.  With the $G$ sampled continuations we form an 
unbiased token-average estimate:
\begin{equation}
D_{\mathrm{KL}}(p)=
\frac{1}{G}\sum_{i=1}^{G}
\frac{1}{|a_i|}\sum_{t=1}^{|a_i|}
\log\left(
\frac{
\pi_\theta(a_{i,t}\mid s_{i,t})
}{
\pi_{\text{ref}}(a_{i,t}\mid s_{i,t})
}
\right),
\label{eq:GRPOKL}
\end{equation}
which measures how far the current policy drifts from the frozen reference 
$\pi_{\text{ref}}$ over the \emph{whole} continuation.

Relative to PPO in Equ.~\eqref{eq:ppo}, GRPO introduces two key improvements: \textit{1) Eliminates the value (critic) network.} Variance reduction is achieved by a \emph{group-relative} baseline, leading to lower memory footprint and fewer hyper-parameters. 
\textit{2) Separates the KL loss channel.} The KL divergence is optimized as an explicit regulariser rather than folded into the advantage, yielding a transparent trade-off between reward maximization and reference anchoring.


%% file: symbol.tex
\begin{tabular}{llll}
\toprule
\textbf{Symbol} & \textbf{Alias} &
\textbf{Meaning} &
\textbf{Appears in}\\
\midrule
$p$ & prompt & User prompt (initial state) &
\S\ref{subsec:notation}, \S\ref{subsec:rlhf},
\S\ref{subsec:ppo}, \S\ref{subsec:grpo}\\

$a_t$ & action & Token / pixel patch / diffusion noise at step $t$ &
\S\ref{subsec:notation}, Eq.~\eqref{eq:ppoKL},
\S\ref{subsec:grpo} \\

$y$ & traj & Full continuation $(a_1,\dots,a_T)$ &
\S\ref{subsec:notation}, \S\ref{subsec:rlhf},
\S\ref{subsec1:rlvr} \\

$y_i$ & continuation & $i$-th continuation in a GRPO group &
\S\ref{subsec:notation}, \S\ref{subsec:grpo} \\

$s_t$ & state & Prompt plus previously generated actions &
\S\ref{subsec:notation}, \S\ref{subsec:ppo},
\S\ref{subsec:grpo} \\

$\pi_\theta$ & policy & Trainable model (current parameters) &
\S\ref{subsec:notation}, \S\ref{subsec:rlhf},
\S\ref{subsec:dpo}, \S\ref{subsec:ppo},
\S\ref{subsec:grpo} \\

$\pi_{\theta_{\mathrm{old}}}$ & behaviour policy &
Frozen policy that produced current batch &
Eq.~\eqref{eq:ppo}, \S\ref{subsec:grpo} \\

$\pi_{\text{SFT}}$ & SFT baseline &
Supervised-fine-tuned checkpoint &
\S\ref{subsec:rlhf}, \S\ref{subsec:ppo} \\

$\pi_{\text{ref}}$ & reference &
Policy used in KL regulariser &
\S\ref{subsec:notation}, \S\ref{subsec:ppo},
\S\ref{subsec:grpo} \\

$\rho_t$ & ratio &
Importance weight $\pi_\theta/\pi_{\theta_{\mathrm{old}}}$ &
Eq.~\eqref{eq:ppo}, \S\ref{subsec:grpo} \\

$V_\psi$ & critic &
Value network predicting future return &
\S\ref{subsec:ppo} \\


$\hat A_t$ & advantage &
GAE advantage (token-level) &
Eq.~\eqref{eq:ppo_GAE}, \S\ref{subsec:ppo} \\

$\hat{A}_{i,t}$ & group adv. &
Group-normalised advantage (GRPO) &
Eq.~\eqref{eq:grpo_obj}, \S\ref{subsec:grpo} \\

$O=\{a_t\}_{1}^{G}$ & group &
Set of $G$ continuations for one prompt at $t$ timestep &
\S\ref{subsec:notation}, \S\ref{subsec:grpo} \\

$G$ & group size & Number of continuations per prompt &
\S\ref{subsec:notation}, \S\ref{subsec:grpo} \\

$r_\phi(s_t,a_t)$ & token reward &
Immediate reward from frozen preference model &
\S\ref{subsec:ppo} \\

$r_i$ & token reward &
Reward of the $i$-th continuation in group &
\S\ref{subsec:grpo} \\

$\operatorname{mean}(\cdot)$ & mean &
Group reward mean in GRPO &
\S\ref{subsec:grpo} \\

$\operatorname{std}(\cdot)$ & std &
Group reward standard deviation in GRPO &
\S\ref{subsec:grpo} \\

$R_\phi(p,y)$ & RM &
Sequence-level reward model (RLHF) &
\S\ref{subsec:rlhf} \\

$\epsilon$ & clip &
PPO clipping threshold &
Eq.~\eqref{eq:ppo}, \S\ref{subsec:grpo} \\

$\beta$ & KL weight &
Weight balancing KL vs.\ reward &
Eq.~\eqref{eq:ppoKL}, \S\ref{subsec:rlhf},
\S\ref{subsec:grpo} \\

$\operatorname{KL}(\!\cdot\!\|\!\cdot\!)$ & KL &
Divergence between policy and reference &
\S\ref{subsec:rlhf}, \S\ref{subsec:ppo},
\S\ref{subsec:grpo} \\

$D_{\mathrm{KL}}(p)$ & est.\ KL &
Token-average KL estimator in GRPO &
Eq.~\eqref{eq:GRPOKL} \\

\bottomrule
\end{tabular}

%% file: 3_main_content.tex
\section{Reinforcement Learning in Vision}
\label{sec:3_visual_reinforcement_learning}

\input{category_tree}

\subsection{Multi-Modal Large Language Models}
\label{Multi_Modal_Large_Language_Models}

We categorize the works into four coherent groups, each defined by shared RL-driven objectives and internal reasoning mechanisms. 
%

\subsubsection{Conventional RL-based MLLMs}\label{subsec:trad_rl_mllm}

We refer to \emph{conventional RL-based MLLMs} as approaches that apply reinforcement learning primarily to \textit{align} a vision–language backbone with \emph{verifiable, task-level rewards}, without explicitly modeling multi-step chain-of-thought reasoning.
Typical works RePIC~\cite{oh2025repic}, GoalLadder~\cite{zakharov2025goalladder}, Drive-R1~\cite{li2025drive} and VLM-R1~\cite{shen2025vlm} replace preference models with \emph{deterministic validators} (\textit{e.g.,} exact-match, IoU, BLEU) and optimize the policy by GRPO/PPO variants under a KL regulariser.
This design yields stable value-free training, improves zero-shot robustness on captioning, grounding and autonomous-driving benchmarks, and substantially reduces the annotation cost typically incurred by supervised fine-tuning.

Recent extensions demonstrate the flexibility of this paradigm.  
GRPO-CARE~\cite{chen2025grpo} introduces consistency-aware group normalization to mitigate reward variance, while Q-Ponder~\cite{cai2025q} adds a pondering controller.
From a data perspective, MoDoMoDo formulates a multi-domain mixture optimization that predicts reward distributions and selects optimal curricula~\cite{liang2025modomodo};
V-Triune further unifies perception and reasoning tasks within a single triple-objective pipeline, empirically validating that rule-based RL scales to diverse visual signals~\cite{ma2025one}.  
Collectively, these studies indicate that (i) verifiable rewards can serve as a low-cost alternative to human feedback, (ii) group-relative objectives offer higher training stability than token-level PPO on heterogeneous visual tasks, and (iii) curriculum or data-mixture scheduling is emerging as a key ingredient for broad generalization.

\subsubsection{Spatial and 3D Perception }
\label{Spatial_Perception}

\textbf{2D perception.}
Perception centric works applies RL to sharpen object detection, segmentation and grounding without engaging in lengthy chain–of–thought reasoning.
Omni-R1~\cite{zhong2025omni} introduces a two-system (global–local) GRPO pipeline that verifies predictions via rule-based metrics, yielding notable improvements on region-level benchmarks and emotion-recognition tasks.
DIP-R1~\cite{park2025dip} further decomposes perception into step-wise “inspect $\!\rightarrow\!$ observe $\!\rightarrow\!$ act” cycles, where each stage receives deterministic IoU or counting-based rewards to boost fine-grained detection.
Perception-R1~\cite{yu2025perception} revisits the effect of GRPO on a spectrum of detection and OCR datasets.
Complementing these, VisRL~\cite{chen2025visrl} frames intention-guided focus selection as an RL sub-policy, eliminating expensive region labels and consistently outperforming supervised strong baselines on visual grounding tasks.

\textbf{3D perception.}
Beyond 2-D, several studies leverage RL to align multimodal models with physically consistent 3-D layouts.  
MetaSpatial~\cite{pan2025metaspatial} employs rendered depth/IoU rewards to refine spatial reasoning for AR/VR scene generation, whereas Scene-R1~\cite{yuan2025scene} couples video-grounded snippet selection with a two-stage grounding policy to learn 3-D scene structure without point-level supervision.
At molecular scale, BindGPT~\cite{zholus2025bindgpt} treats atom placement as sequential actions and uses binding-affinity estimators as verifiable rewards, demonstrating the scalability of perception-focused RL to 3-D biochemical design.
Collectively, these approaches underscore a common recipe: (i) formulate detection/segmentation/3-D alignment as Markov decision problems, (ii) craft deterministic spatial rewards (\textit{e.g.,} IoU, depth consistency, binding energy), and (iii) fine-tune pretrained VLM backbones via GRPO/PPO for stable perception enhancement—thereby differentiating themselves from reasoning-oriented RL variants.

\subsubsection{Image Reasoning}
\label{Spatial_Reasoning}

Thinking about Images methods enhance multimodal reasoning by
\emph{verbalising} observations of a static picture before producing an answer,
but the visual content itself is not modified during inference. 
By contrast, Thinking with Images elevates the picture to an \emph{active,
external workspace}: models iteratively \emph{generate, crop, highlight, sketch
or insert explicit visual annotations} as tokens in their chain-of-thought,
thereby aligning linguistic logic with grounded visual evidence.
%

\textbf{Think about Image}.
\label{Think_about_Vision}
Early think about image works for spatial VQA employs view-consistent or transformation-invariant objectives, such as SVQA-R1~\cite{wang2025svqa} and STAR-R1~\cite{li2025star}.
VL-GenRM~\cite{zhang2025vl} and RACRO~\cite{gou2025perceptual} refine preference data or caption rewards to curb hallucinations.
Benchmark-oriented efforts such as EasyARC~\cite{unsal2025easyarc} offer procedurally generated, fully verifiable tasks that suit outcome-based reinforcement learning.
To mitigate shortcut reliance and improve generality, Visionary-R1~\cite{xia2025visionary} enforces image interpretation before reasoning, whereas UniVG-R1~\cite{bai2025univg} unifies referring, captioning, and detection by coupling a grounding corpus with GRPO fine-tuning. 
Extensions such as EchoInk-R1~\cite{xing2025echoink} further enrich visual reasoning by integrating audio–visual synchrony under GRPO optimization.
Meanwhile, curriculum-driven frameworks—WeThink, G1, GThinker, and Observe-R1 progressively increase task complexity or introduce re-thinking cues (\textit{e.g.,} difficulty ladders, multimodal format constraints) to cultivate deeper and more structured reasoning capabilities in MLLMs.
These methods show that language only RL with well-designed visual correctness rewards significantly improves model accuracy, robustness, and out-of-distribution performance.

\textbf{Think with Image.}
\label{Think_with_Vision}
Early think with image works grounds reasoning via discrete \textit{region–level} operations:
GRIT~\cite{fan2025grit} interleaves bounding-box tokens with language and trains under GRPO to maximize both answer correctness and box fidelity, while VILASR~\cite{wu2025reinforcing} generalizes this idea to multi-view and video settings, enforcing cross-view spatial consistency.
Ground-R1~\cite{cao2025ground} and BRPO~\cite{chu2025qwen} adopt two-stage pipelines that first highlight evidence regions (via IoU-based or reflection rewards) before verbal reasoning.
A complementary thread explores pixel-space or sequence-level manipulation.
Visual Planning~\cite{xu2025visual} replaces text chains with imagined image roll-outs rewarded by downstream task success;
Pixel Reasoner~\cite{su2025pixel} augments the action space with crop, erase and paint primitives and balances exploration through curiosity-driven rewards, whereas DeepEyes~\cite{zheng2025deepeyes} shows that end-to-end RL can spontaneously induce such visual thinking behaviours.
Finally, TACO~\cite{kan2025taco} introduces a think–answer consistency objective that resamples long visual–verbal chains until their intermediate edits align with the final answer. 
Together, these systems demonstrate that explicitly generating or editing visual artefacts during reasoning optimized via GRPO or R1-style outcome RL—yields more faithful, interpretable and robust image understanding than language only approaches.

\subsubsection{Video Reasoning}
\label{Video_Reasoning}

Video reasoning extends the capabilities of MLLMs to process temporal dynamics, requiring not only spatial perception but also sequential understanding and causal inference.
Recent works in this domain have proposed diverse approaches to tackle complex reasoning over video inputs.
For instance, VQ-Insight~\cite{zhang2025vq} introduces a hierarchical reward design and self-consistency voting mechanism tailored to the question–answering process over long videos. 
TW-GRPO~\cite{dang2025reinforcing} combines token wise credit assignment with GRPO-style optimization to improve fine-grained temporal alignment between textual reasoning and video evidence.
Meanwhile, several R1-style frameworks have been developed to unlock video understanding in complex real-world or egocentric settings. EgoVLM~\cite{vinod2025egovlm} and VAU-R1~\cite{zhu2025vau} focus on egocentric video reasoning with visual memory and utility-based rewards.
DeepVideo-R~\cite{park2025deepvideo} integrates dense video encodings and external reward functions to supervise long-horizon reasoning. 
TimeMaster~\cite{zhang2025timemaster} explicitly structures temporal abstraction and reasoning via curriculum learning, while VideoR1~\cite{videor1} proposes a scalable RL framework for video-based QA tasks across multiple domains.
Collectively, these works highlight the importance of aligning temporal representations with language trajectories through reinforcement learning, paving the way for robust and generalizable video reasoning agents.

\subsection{Visual Generation}
\label{sec:Visual_Generation}

\subsubsection{Image Generation}
\label{image_gen}

RL for image generation models departs from the language counterpart in both \emph{action space} continuous diffusion steps or prompt refinements and \emph{reward design}, which must capture perceptual quality, text–image alignment, and subject fidelity.
A first family learns an explicit \emph{visual reward model}: ImageReward~\cite{xu2023imagereward} supplies human-preference scores that drive policy-gradient fine-tuning of diffusion backbones in DiffPPO~\cite{xiao2024diffppo}, Dpok~\cite{fan2023dpok}, and FocusDiff~\cite{pan2025focusdiff}.
A second line bypasses reward modelling by optimising \emph{pairwise or unary preferences}: DDPO~\cite{black2023ddpo}, DiffusionDPO~\cite{wallace2024diffusion}, Diffusion-KTO~\cite{li2024aligning}, and DiffusionRPO~\cite{gu2024diffusion} treat denoising trajectories as MDPs and apply R1/GRPO updates to maximise comparative human feedback. 
Beyond alignment, works such as PrefPaint~\cite{liu2024prefpaint}, Parrot~\cite{lee2024parrot}, and RLRF~\cite{rodriguez2025rendering} craft multi-objective or render-and-compare rewards to refine aesthetics, diversity, or vector graphics.
RL has also been used to inject \emph{reasoning and prompt adaptation}: ReasonGen-R1~\cite{zhang2025reasongen}, GoT-R1~\cite{duan2025got} and RePrompt~\cite{wu2025reprompt} first generate textual plans or improved prompts, then reinforce the generator for coherent scene synthesis.
Finally, personalisation methods—DPG-T2I~\cite{wei2024powerful}, RPO~\cite{miao2024subject}, and B$^{2}$-DiffuRL~\cite{hu2025towards} optimize rewards that measure subject fidelity under scarce reference images. 
Collectively, these studies show that RL, armed with perceptual or preference-based rewards, can steer diffusion models toward higher realism, stronger prompt faithfulness, controllable layout, and user-specific appearance capabilities difficult to achieve with likelihood training alone.

\subsubsection{Video Generation}
\label{video_gen}

Applying RL to video generation introduces challenges absent in images: rewards must capture \emph{temporal coherence}, \emph{motion naturalness}, and \emph{text–video alignment} across hundreds of frames.
Early work such as InstructVideo~\cite{yuan2024instructvideo} repurposed image scorers and applied PPO to refine short clips, whereas VideoRM~\cite{wu2024boosting} and VideoReward~\cite{liu2025improving} learn dedicated preference models that grade entire sequences for smoothness, consistency and prompt faithfulness.  Building on GRPO/R1, DanceGRPO~\cite{xue2025dancegrpo} shows that group-normalized returns stabilize long-horizon optimization and boost aesthetic and alignment scores across diffusion and flow backbones.  

Beyond preference alignment, specialized rewards tackle domain-specific goals.  
GAPO~\cite{zhu2025aligning} exploits gap-aware ranking to fine-tune anime videos; Phys-AR~\cite{lin2025reasoning} penalizes violations of physics to yield plausible trajectories; and InfLVG~\cite{fang2025inflvg} trains an inference-time policy that retains only context tokens beneficial for multi-scene coherence.
Auxiliary critics further facilitate training: VideoScore~\cite{he2024videoscore} and Gradeo~\cite{mou2025gradeo} offer explainable, multi-factor scores, while TeViR~\cite{chen2025tevir} feeds imagined video roll-outs as dense rewards to downstream control agents.
Collectively, these studies demonstrate that carefully crafted sequence level rewards and group-relative policy updates are key to producing temporally consistent, semantically aligned, and physically plausible videos capabilities difficult to obtain with likelihood training alone.

\subsubsection{3D Generation}
\label{3D_gen}

RL for 3D generation differs from 2D and video tasks as rewards operate on \emph{volumetric structures} or \emph{rendered views}, often requiring expensive render-and-compare loops.  
DreamCS~\cite{zou2025dreamcs} pioneers this paradigm by framing text-to-mesh synthesis as a Markov decision process: a diffusion backbone proposes coarse shapes, then a policy refines vertex positions under a reward that jointly measures silhouette IoU, CLIP text‐mesh alignment, and mesh smoothness.
On the reward side, DreamReward~\cite{ye2024dreamreward} introduces a large‐scale human preference dataset of rendered 3-D assets and trains a geometry-aware critic that scores prompts, renders, and latent SDFs; the learned signal enables stable PPO fine-tuning of existing text-to-shape models.

A complementary line adopts direct preference optimization. 
DreamDPO~\cite{zhou2025dreamdpo} extends DPO to NeRF and mesh diffusion models by sampling paired 3D outputs and maximizing the margin dictated by human comparisons, achieving superior prompt fidelity without an explicit value network.
Finally, Nabla-R2D3~\cite{liu2025nabla} tackles \emph{interactive scene editing}: the agent sequentially adds, deletes, or transforms objects in a 3D scene; reward is computed via real-time rasterized views and task-specific validators (\eg occupancy, spatial relations).  Group-relative policy optimization (R2D3) stabilizes training despite sparse feedback. 
Together, these studies demonstrate that RL, equipped with geometry-aware or render-based rewards, provides an effective mechanism for controlling structural integrity, text alignment, and interactive editability capabilities that standard likelihood or score-distillation objectives struggle to capture in the 3D domain.

\begin{figure*}[t]
    \centering
	\includegraphics[width=0.9\linewidth]{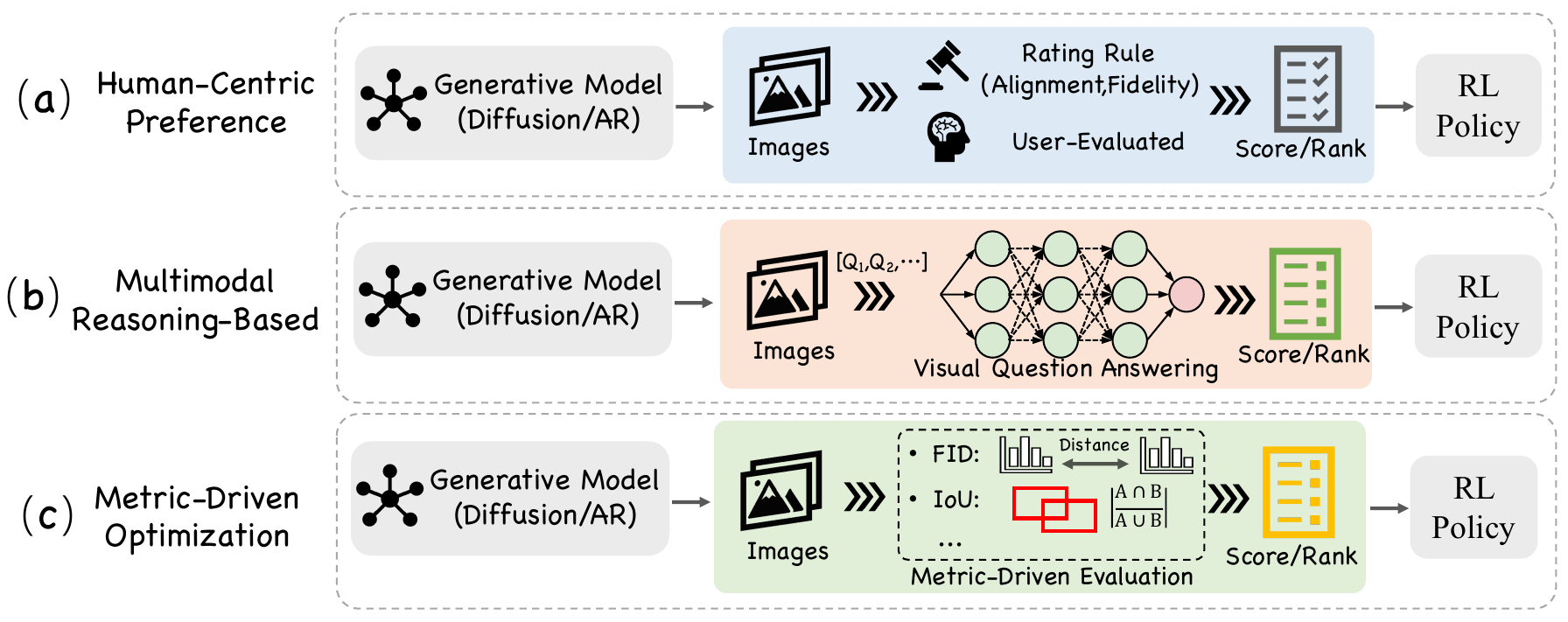}
	\caption{\textbf{Three reward paradigms for RL-based image generation.} (a) Human-Centric Preference Optimization: aligns outputs with human aesthetic scores (HPS~\cite{wu2023human}, ImageReward~\cite{xu2023imagereward}); (b) Multimodal Reasoning-Based Evaluation: scores images via multimodal reasoning consistency (UnifiedReward~\cite{wang2025unified}, PARM~\cite{guo2025can}); (c) Metric-Driven Objective Optimization: minimizes task-specific quantitative metrics such as FID and IoU.
    }
\label{reward}
\end{figure*}

\subsubsection{Reward Design and Preference Modeling for Visual Generation}

As shown in Figure~\ref{reward}, the reward design in RL-based visual generation can broadly be divided into three paradigms: 
(a) Human-Centric Preference Optimization (using human feedback or learned reward models for video alignment and fidelity). 
Representative works include ImageReward~\cite{xu2023imagereward}, which pioneered large-scale text-to-image reward modeling from human comparisons and provided a critic for diffusion fine-tuning.
Dpok~\cite{fan2023dpok} and DiffPPO~\cite{xiao2024diffppo} adapted PPO with KL regularization and human aesthetic scores to enhance alignment and fidelity.
Extending these ideas to video, VideoRM~\cite{wu2024boosting} and VideoReward~\cite{liu2025improving} trained reward models on large-scale human (and GPT-assisted) video comparisons, enabling RL fine-tuning of text-to-video generation.
(b) Multimodal Reasoning-Based Evaluation (using a question-answering or reasoning module to score consistency between videos and textual descriptions).
For example, LLaVA-Reward~\cite{zhou2025multimodalllmscustomizedreward} leverages a large VLM to evaluate images across multiple dimensions including alignment, fidelity, and safety, providing a multi-faceted reward signal that better approximates human judgment.
and (c) Metric-Driven Objective Optimization (directly optimizing quantitative metrics like FID or IoU for quality).

\subsection{Unified Model}
\label{sec:Unified_Model}

Task specific RL maximizes a reward tied to a \emph{single} objective, whereas \emph{Unified RL} optimizes a \emph{shared} policy and reward across multiple vision–language tasks (\eg~understanding and generation).

\subsubsection{Unified RL}
\label{Unified_RL}

Unlike task specific pipelines that attach RL to a single downstream objective, \emph{Unified RL} methods optimize a \textbf{shared policy} across heterogeneous multimodal tasks under a \textit{single} reinforcement signal.
The central idea is to merge understanding and generation trajectories into one training loop typically using Group-Relative or R1-style methods.

UniRL~\cite{mao2025unirl} exemplifies this paradigm: a visual autoregressive backbone is first instruction-tuned, then jointly fine-tuned on VQA, captioning and image generation with a blended reward measuring textual correctness, CLIP-based alignment, and aesthetic quality.
CoRL~\cite{jiang2025co} pushes the idea further by alternating “co-understanding” and “co-generation” batches within the same GRPO step.
To address inefficiency in dense token spaces, SelfTok~\cite{wang2025discrete} discretises multi-modal actions into a self-evolving token set and demonstrates that a single RL head can govern retrieval, grounding, and synthesis with minimal extra parameters.
Finally, HermesFlow~\cite{yang2025hermesflow} couples an autoregressive text module with a rectified flow image decoder under one cross-task reward, illustrating that diffusion-style and language-style policies can be harmonized through unified reinforcement updates.  
Together, these works suggest that sharing a common RL objective across tasks not only reduces training cost but also encourages emergent cross modal generalization unavailable to isolated, task specific fine-tuning.

\subsubsection{Task specific RL}
\label{Task_specific_RL}

In contrast to the unified approaches of \S \ref{Unified_RL}, task-specific RL confines the reward signal to a single downstream objective, optimizing one functional head while leaving other capabilities untouched.
VARGPT-v1.1~\cite{zhuang2025vargpt} exemplifies this strategy: although the underlying visual autoregressive model can handle both understanding and generation, its RL stage targets \emph{only} visual generation with DPO.
Similarly, Emu3~\cite{wang2024emu3} introduces RL exclusively to polish its image generation branch, which leveraging pair wise human preferences.
For the multimodal understanding abilities of model (\eg~captioning, VQA), the work just train this part by task specific fine-tuning alone.

\subsection{Vision Language Action Models}
\label{sec:Vision_Language_Action_Models}

\subsubsection{GUI Automation}
\label{GUI_Interaction_Agents}

Modern GUI RL research frames screen understanding and action prediction as a vision–language decision process, then employs rule-based or preference rewards to close the perception–action loop.
On desktop and web interfaces, GUI-R1~\cite{luo2025gui} introduces an R1-style rule set that maps click success, text entry, and scroll validity to dense rewards.
UI-R1~\cite{lu2025ui} adds GRPO with a novel action-specific KL term to stabilize long-horizon plans, while SE-GUI~\cite{yuan2025enhancing} applies self-evolutionary filtering to distil high-fidelity trajectories.
Focusing on trajectory reuse, UIShift~\cite{gao2025uishift} formulates an inverse dynamics objective that lets MLLM learn actions from unlabeled GUI pairs and then refines them via RL.
Complementary preference-based frameworks include LPO~\cite{tang2025lpo} that rewards spatial proximity for precise clicks.
ProgRM~\cite{zhang2025progrm} injects program-level logical checks, and RUIG~\cite{zhang2023reinforced} leverages instruction grounding with reinforcement signals.
Tool-specific baselines such as Ui-tars~\cite{qin2025ui} offer larger action vocabularies yet still rely on rule-driven RL for robust execution.

Mobile scenarios introduce latency and on-device constraints.
AgentCPM-GUI~\cite{zhang2025agentcpm} compresses the action space and conducts GRPO fine-tuning.
MobileGUI-RL~\cite{shi2025mobilegui} advances this line via online RL with task-level rewards to improve exploration under limited memory,
and Mobile-R1~\cite{gu2025mobile} extends interactive multi-turn RL to correct error cascades during long tasks.
At inference, GTA1~\cite{yang2025gta1} samples multiple action candidates and employs a judge model to pick the best, effectively trading compute for higher success rates.
Additional, light-weight models such as Appvlm~\cite{papoudakis2025appvlm} demonstrate that modest-sized MLLM, after GRPO fine-tuning, can control smartphone apps with competitive precision.
To adaptively reduce the thinking length, the TON~\cite{ton} proposes a thought-dropout solution during supervised fine-tuning stage, then GRPO skill adaptivley skip unnecessary reasoning process for efficiently thinking.

Collectively, these studies show that GUI agents benefit from rule-verifiable rewards, group-normalzsed policy updates, and preference-guided localization, achieving rapid progress toward reliable, cross-platform automation.

\subsubsection{Visual Navigation}
\label{Vision_Navigation_Robotics}

RL‐driven visual navigation research now couples large vision–language models with embodied control, employing group‐normalized or time‐decayed returns to maintain long‐horizon stability.
OctoNav-R1~\cite{gao2025octonav} exploits a hybrid RL pipeline with a ``think-before-action'' ability for VLA model, then translate egocentric frames into low-level actions. 
Focusing on dataset efficiency, VLN-R1~\cite{qi2025vln} builds an end-to-end navigator and introduces a time-decayed reward to handle continuous trajectories.
At the system level, Flare~\cite{hu2024flare} demonstrates that fine-tuning a multi-task robotics policy with large-scale RL in simulation can generalize to real-world household tasks.
Complementary advances include More~\cite{zhao2025more}, which augments omni-directional inputs with memory-guided policy distillation, and RAPID~\cite{kim2025rapid}, which integrates pose priors for faster convergence in unseen layouts.
These works show that using temporal rewards, memory sampling, and environment priors with GRPO/PPO helps VLA agents navigate more reliably and efficiently.

\subsubsection{Visual Manipulation}
\label{Dexterous_Manipulation_Robotics}

Visual manipulation tasks (\eg~object relocation, tool use, and multi-step rearrangement) require fine-grained perception and long-horizon planning.
Recent works~\cite{chen2025tgrpo,shu2025rftf} integrate reinforcement learning with vision–language models to enhance generalization, interactivity, and policy consistency.
TGRPO~\cite{chen2025tgrpo} introduces a task-grounded reward formulation and group normalized updates to stabilize training for open-ended object manipulation.
RFTF~\cite{shu2025rftf} applies rule-based rewards to support interactive table top tasks and emphasizes training with minimal human supervision.
Meanwhile, RLVLA~\cite{liu2025can} and VLA-RL~\cite{lu2025vla} explore curriculum-based or progressive reinforcement learning for VLM-based robot agents, achieving high success rates across diverse rearrangement environments.

Building on this, ConRFT~\cite{chen2025conrft} and iRe-VLA~\cite{guo2025improving} introduce consistency-aware and instruction-refinement strategies respectively, using RL to align visual predictions with physical interaction outcomes.
RIPT-VLA focuses on interactive prompting during manipulation, bridging LLM planning and low-level control through reinforced feedback~\cite{tan2025interactive}. 
Finally, ReinBot~\cite{zhang2025reinbot} leverages multimodal rollouts and preference-based updates to improve real-world manipulation robustness~\cite{zhang2025reinbot}. 
Collectively, these studies highlight the role of vision-language reasoning, structured reward design, and RL-based refinement in advancing embodied manipulation under complex, language-conditioned settings.

%% file: category_tree.tex
\tikzstyle{my-box}=[
    rectangle,
    rounded corners,
    text opacity=1,
    minimum height=1.5em,
    minimum width=5em,
    inner sep=2pt,
    align=center,
    fill opacity=.5,
]
\tikzstyle{cause_leaf}=[my-box, minimum height=1.5em,
    fill=lighttealblue!20, text=black, align=left,font=\scriptsize,
    inner xsep=2pt,
    inner ysep=4pt,
]
\tikzstyle{detect_lightcoral}=[my-box, minimum height=1.5em,
    fill=lightcoral!20, text=black, align=left,font=\scriptsize,
    inner xsep=2pt,
    inner ysep=4pt,
]

\tikzstyle{detect_burlywood}=[my-box, minimum height=1.5em,
    fill=burlywood!20, text=black, align=left,font=\scriptsize,
    inner xsep=2pt,
    inner ysep=4pt,
]
\tikzstyle{detect_leaf}=[my-box, minimum height=1.5em,
    fill=lightplum!20, text=black, align=left,font=\scriptsize,
    inner xsep=2pt,
    inner ysep=4pt,
]

\tikzstyle{detect_lightsteelblue}=[my-box, minimum height=1.5em,
    fill=lightsteelblue!20, text=black, align=left,font=\scriptsize,
    inner xsep=2pt,
    inner ysep=4pt,
]
\tikzstyle{detect_wheat}=[my-box, minimum height=1.5em,
    fill=wheat!20, text=black, align=left,font=\scriptsize,
    inner xsep=2pt,
    inner ysep=4pt,
]
\tikzstyle{detect_tan}=[my-box, minimum height=1.5em,
    fill=tan!20, text=black, align=left,font=\scriptsize,
    inner xsep=2pt,
    inner ysep=4pt,
]
\tikzstyle{detect_lavender}=[my-box, minimum height=1.5em,
    fill=lavender!20, text=black, align=left,font=\scriptsize,
    inner xsep=2pt,
    inner ysep=4pt,
]
\tikzstyle{detect_harvestgold}=[my-box, minimum height=1.5em,
    fill=harvestgold!20, text=black, align=left,font=\scriptsize,
    inner xsep=2pt,
    inner ysep=4pt,
]
\begin{figure*}[t]
    \centering
    \resizebox{\textwidth}{!}{
        \begin{forest}
            forked edges,
            for tree={
                grow=east,
                reversed=true,
                anchor=base west,
                parent anchor=east,
                child anchor=west,
                base=left,
                font=\scriptsize,
                rectangle,
                rounded corners,
                align=left,
                minimum width=4em,
                edge+={darkgray, line width=1pt},
                s sep=3pt,
                inner xsep=2pt,
                inner ysep=3pt,
                ver/.style={rotate=90, child anchor=north, parent anchor=south, anchor=center},
            },
            where level=1{text width=8.8em,font=\footnotesize,}{},
            where level=2{text width=10.2em,font=\footnotesize,}{},
            where level=3{text width=7.2em,font=\footnotesize,}{},
            where level=4{text width=35.5em,font=\footnotesize,}{},
            [
                Reinforcement Learning in Vision, ver, color=carminepink!100, fill=carminepink!15,
                text=black
                [
                    Multimodal Large \\
                    Language Models (\S \ref{Multi_Modal_Large_Language_Models}), color=lighttealblue!100, fill=lighttealblue!100, text=black
                    [
                        Conventional RL-based \\Frameworks(\S \ref{subsec:trad_rl_mllm}), color=lighttealblue!100, fill=lighttealblue!60, text=black
                        [
                            {\eg~RePIC~\cite{oh2025repic}, GoalLadder~\cite{zakharov2025goalladder}, Drive-R1~\cite{li2025drive}, VLM-R1~\cite{shen2025vlm}, GRPO-CARE~\cite{chen2025grpo}, Q-Ponder~\cite{cai2025q}, MoDoMoDo~\cite{liang2025modomodo},\\ V-Triune~\cite{ma2025one},Vision-R1~\cite{zhan2025vision}, ProxyThinker~\cite{xiao2025proxythinker}, Jigsaw-R1~\cite{wang2025jigsaw}, SRPO~\cite{wan2025srpo}, R1-Onevision~\cite{yang2025r1}, Reason-RFT~\cite{tan2025reason},\\ Segagent~\cite{zhu2025segagent},
                            VisualQuality-R1~\cite{wu2025visualquality}, Zhang et al.~\cite{zhang2025improving}, SATORI-R1~\cite{shen2025satori}, Skywork r1v2~\cite{wang2025skywork}, Yang et al.~\cite{yang2025self}, THOR~\cite{chang2025thor}\\ Caprl~\cite{xing2025caprl}, LENS~\cite{zhu2025lens}, Docr1~\cite{xiong2025docr1}, Bigcharts-r1~\cite{masry2025bigcharts}, \textit{etc}.}
                                , cause_leaf, text width=38em
                        ]
                    ]
                    [
                        Spatial \& 3D Perception \\(\S \ref{Spatial_Perception}), color=lighttealblue!100, fill=lighttealblue!60,  text=black
                        [
                            {\eg~Omni-R1~\cite{zhong2025omni}, DIP-R1~\cite{park2025dip}, Metaspatial~\cite{pan2025metaspatial},   Bindgpt~\cite{zholus2025bindgpt}, Scene-R1~\cite{yuan2025scene}, Perception-R1~\cite{yu2025perception}, VisRL~\cite{chen2025visrl},\\  ViCrit~\cite{wang2025vicrit}, VisionReasoner~\cite{liu2025visionreasoner}, ViGoRL~\cite{sarch2025grounded}, Visual-RFT~\cite{liu2025visual}, \textit{etc}.}
                            , cause_leaf, text width=38em
                        ]
                    ]
                    [
                        Image Reasoning (\S \ref{Spatial_Reasoning}), color=lighttealblue!100, fill=lighttealblue!60,  text=black
                        [
                            Think about Image, color=lighttealblue!100, fill=lighttealblue!40, text=black
                            [
                            {\eg~SVQA-R1~\cite{wang2025svqa}, VL-GenRM~\cite{zhang2025vl}, RACRO~\cite{gou2025perceptual}, EasyARC~\cite{unsal2025easyarc}, STAR-R1~\cite{li2025star}, 
                            Visionary-R1~\cite{xia2025visionary}, UniVG-R1~\cite{bai2025univg}\\, EchoInk-R1~\cite{xing2025echoink}, WeThink~\cite{yang2025wethink}, G1~\cite{chen2025g1}, GThinker~\cite{zhan2025gthinker}, Observe-R1~\cite{guo2025observe}, Mm-eureka~\cite{meng2025mm},  Perception-R1~\cite{xiao2025advancing}, \\MiMo-VL~\cite{coreteam2025mimovltechnicalreport}, SearchExpert~\cite{li2025enhancing}, \textit{etc}.}
                            , cause_leaf, text width=38em
                            ]
                        ]
                        [
                            Think with Image, color=lighttealblue!100, fill=lighttealblue!40, text=black
                            [
                            {\eg~Visual Planning~\cite{xu2025visual}, GRIT~\cite{fan2025grit}, VILASR~\cite{wu2025reinforcing}, BRPO~\cite{chu2025qwen}, Ground-R1~\cite{cao2025ground}, Pixel Reasoner~\cite{su2025pixel}, DeepEyes~\cite{zheng2025deepeyes}\\, TACO~\cite{kan2025taco}, VRAG-RL~\cite{wang2025vrag}, VisTA~\cite{huang2025visualtoolagent},TGI~\cite{chern2025thinking}, Chain-of-Focus~\cite{zhang2025chain}, Openthinkimg~\cite{su2025openthinkimg}, Active-O3~\cite{zhu2025active}\\, RRVF~\cite{chen2025learning}, Visionthink~\cite{yang2025visionthink}, \textit{etc}.}
                            , cause_leaf, text width=38em
                            ]
                        ]
                    ]
                    [
                        Video Reasoning(\S \ref{Video_Reasoning}), color=lighttealblue!100, fill=lighttealblue!60, text=black
                        [
                            {\eg~VQ-Insight~\cite{zhang2025vq}, TW-GRPO~\cite{dang2025reinforcing}, EgoVLM~\cite{vinod2025egovlm}, VAU-R1~\cite{zhu2025vau}, DeepVideo-R~\cite{park2025deepvideo}, TimeMaster~\cite{zhang2025timemaster},\\ VideoR1~\cite{videor1}, VideoChat-R1~\cite{li2025videochat}, Temporal-RLT~\cite{li2025reinforcement}, Moss-chatv~\cite{tao2025moss}, Tempsamp-r1~\cite{li2025tempsamp}, Video-mtr~\cite{xie2025video},\\ Thinking with videos~\cite{zhang2025thinking}, MIRO~\cite{dufour2025miro}, Chunk-GRPO~\cite{luo2025sample}, Onereward~\cite{gong2025onereward}, Uso~\cite{wu2025uso}, Pref-grpo~\cite{wang2025pref}\\, Tempflow-grpo~\cite{he2025tempflow},  \textit{etc}.}
                                , cause_leaf, text width=38em
                        ]
                    ]
                ]
                [
                    Visual Generation(\S \ref{sec:Visual_Generation}), color=lightplum!100, fill=lightplum!100, text=black
                    [   
                        Image Generation(\S \ref{image_gen}), color=lightplum!100, fill=lightplum!60, text=black
                        [
                            {\eg~ImageReward~\cite{xu2023imagereward},  ReasonGen-R1~\cite{zhang2025reasongen}, FocusDiff~\cite{pan2025focusdiff}, Dpok~\cite{fan2023dpok}, LOOP~\cite{gupta2025simple}, Prefpaint~\cite{liu2024prefpaint}, RLRF~\cite{rodriguez2025rendering}\\, GoT-R1~\cite{duan2025got}, D-Fusion~\cite{hu2025d}, Black et al.,~\cite{black2023ddpo}, DiffusionDPO~\cite{wallace2024diffusion}, Diffusion-KTO~\cite{li2024aligning},DiffusionRPO~\cite{gu2024diffusion},\\ Miao et al.,~\cite{miao2024training}, Parrot~\cite{lee2024parrot}, DPG-T2I~\cite{wei2024powerful}, RPO~\cite{miao2024subject}, RePrompt~\cite{wu2025reprompt}, RLCM~\cite{oertell2024rl}, B2-DiffuRL~\cite{hu2025towards},\\ DiffPPO~\cite{xiao2024diffppo}, Simplear\cite{wang2025simplear}, LLaVA-Reward~\cite{zhou2025multimodalllmscustomizedreward}, Flow-grpo~\cite{liu2025flow}, Gallici et al.~\cite{gallici2025fine}, TexForce~\cite{chen2024enhancing}, \textit{etc}.}
                                , detect_leaf, text width=38em
                        ]
                    ]
                    [   
                        Video Generation(\S \ref{video_gen}), color=lightplum!100, fill=lightplum!60, text=black
                        [
                                {\eg~DanceGRPO~\cite{xue2025dancegrpo}, InfLVG~\cite{fang2025inflvg}, Phys-AR~\cite{lin2025reasoning},  VideoReward~\cite{liu2025improving}, TeViR~\cite{chen2025tevir}, GAPO~\cite{zhu2025aligning},\\  Instructvideo~\cite{yuan2024instructvideo}, Videoscore~\cite{he2024videoscore}, Gradeo~\cite{mou2025gradeo}, VideoRM~\cite{wu2024boosting}, \textit{etc}.}
                                , detect_leaf, text width=38em
                        ]
                    ]
                    [
                        3D Generation(\S \ref{3D_gen}), color=lightplum!100, fill=lightplum!60, text=black
                        [
                                {\eg~DreamCS~\cite{zou2025dreamcs}, Dreamreward~\cite{ye2024dreamreward}, DreamDPO~\cite{zhou2025dreamdpo}, Nabla-R2D3~\cite{liu2025nabla}, \textit{etc}.}
                                , detect_leaf, text width=38em
                        ]
                    ]
                ]
                [
                    Unified Model(\S \ref{sec:Unified_Model}), color=harvestgold!100, fill=harvestgold!100, text=black
                    [   
                        Unified RL(\S \ref{Unified_RL}), color=harvestgold!100, fill=harvestgold!60, text=black
                        [
                                {\eg~UniRL~\cite{mao2025unirl}, CoRL~\cite{jiang2025co}, SelfTok~\cite{wang2025discrete}, Hermesflow~\cite{yang2025hermesflow}, UnifiedReward~\cite{wang2025unified}, UnifiedReward-Think~\cite{wang2025unified1}, \textit{etc}.}
                                , detect_harvestgold, text width=38em
                        ]
                    ]
                    [   
                        Task-specific RL(\S \ref{Task_specific_RL}), color=harvestgold!100, fill=harvestgold!60, text=black
                        [
                                {\eg~Vargpt-v1. 1~\cite{zhuang2025vargpt}, Emu3~\cite{wang2024emu3}, X-Omni~\cite{geng2025x}, \textit{etc}.}
                                , detect_harvestgold, text width=38em
                        ]
                    ]
                ]
                [
                    Vision Language Action\\ Models(\S \ref{sec:Vision_Language_Action_Models}), color=burlywood!100, fill=burlywood!100, text=black
                    [   
                        GUI Interaction(\S \ref{GUI_Interaction_Agents}), color=burlywood!100, fill=burlywood!60, text=black
                        [
                                {\eg~GUI-R1~\cite{luo2025gui}, SE-GUI~\cite{yuan2025enhancing}, UI-R1~\cite{lu2025ui}, UIShift~\cite{gao2025uishift}, AgentCPM-GUI~\cite{zhang2025agentcpm}, MobileGUI-RL~\cite{shi2025mobilegui}, ProgRM~\cite{zhang2025progrm}\\, Mobile-R1~\cite{gu2025mobile}, GTA1~\cite{yang2025gta1}, LPO~\cite{tang2025lpo}, Ui-tars~\cite{qin2025ui}, RUIG~\cite{zhang2023reinforced}, Appvlm~\cite{papoudakis2025appvlm}, DigiRL~\cite{bai2024digirl}, ARPO~\cite{arpo}, \textit{etc}.}
                                , detect_burlywood, text width=38em
                        ]
                    ]
                    [   
                        Visual Navigation(\S \ref{Vision_Navigation_Robotics}), color=burlywood!100, fill=burlywood!60, text=black
                        [
                                {\eg~OctoNav-R1~\cite{gao2025octonav}, More~\cite{zhao2025more}, RAPID~\cite{kim2025rapid}, VLN-R1~\cite{qi2025vln}, Flare~\cite{hu2024flare}, IRL-VLA~\cite{jiang2025irl}, Embodied-R~\cite{zhao2025embodied},\\ EmbCLIP-Agent~\cite{eftekhar2023selective}, \textit{etc}.}
                                , detect_burlywood, text width=38em
                        ]
                    ]
                    [
                        Visual Manipulation(\S \ref{Dexterous_Manipulation_Robotics}), color=burlywood!100, fill=burlywood!60, text=black
                        [
                                {\eg~TGRPO~\cite{chen2025tgrpo}, RFTF~\cite{shu2025rftf}, RLVLA~\cite{liu2025can}, VLA-RL~\cite{lu2025vla},  ConRFT~\cite{chen2025conrft}, iRe-VLA~\cite{guo2025improving}, RIPT-VLA~\cite{tan2025interactive}, \\ReinBot~\cite{zhang2025reinbot}, VIKI-R~\cite{kang2025viki}, Robot-R1~\cite{kim2025robot}, RLDG~\cite{xu2024rldg}, HIL-SERL~\cite{luo2024precise}, Modem~\cite{hansen2022modem}, VPG~\cite{zeng2018learning}, \textit{etc}.}
                                , detect_burlywood, text width=38em
                        ]
                    ]
                ]
            ]
        \end{forest}
    }
    \caption{\textbf{Overall taxonomy of reinforcement-learning research in vision.} 
    The chart groups existing work by high-level domain (MLLMs, visual generation, unified models, and vision-language action agents) and then by finer-grained tasks, illustrating representative papers for each branch.}
    \label{fig:categorization_of_survey}
    \vspace{-0.5cm}
\end{figure*}

%% file: 4_benchmark_and_metric.tex
\begin{figure}[t]
    \centering
	\includegraphics[width=0.98\linewidth]{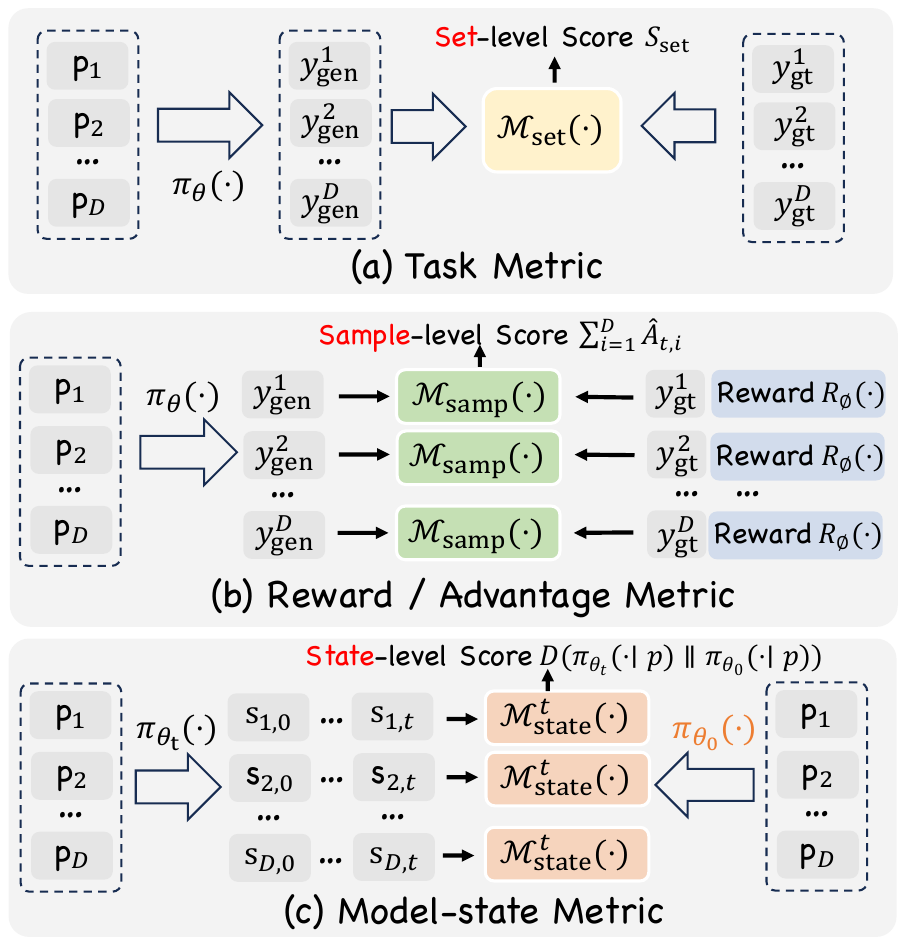}
	\caption{\textbf{Metric Granularity in Visual RL.}
     (a) Set-level metric $\Mset$: one score over the whole prompt set, used for final evaluation (\eg~FID).
    (b) Sample-level metric $\Msamp$: per-output rewards that train the policy (RLHF, DPO).
    (c) State-level metric $\Mstate^t$: training-time signals like KL or length drift, used to monitor stability.
    Notation: $p_i$, $y^i_{\text{gen}}$, $y^i_{\text{gt}}$ denote the prompt, the generated output, and ground truth, respectively.
    $\pi_{\theta_0}$ and $\pi_{\theta_t}$ refer to the $0$-th and $t$-th policy model.
    $R_\phi(\cdot)$ denotes the reward model.
    }
\label{fig:metric}
\end{figure}

\begin{table*}[t]
    \centering
      \scriptsize
      \caption{\textbf{Overview of evaluation metrics in large-model visual RL.}  
               Each task family is broken down into Task Metrics
               (RL-free external benchmarks), Reward Metrics
               (how the learning signal is computed), and
               Model-state Metrics (diagnostics tracked during optimization).}
      \label{tab:metrics_visual_rl}
    \input{table_metric}
\end{table*}

\section{Metrics and Benchmarks}
\label{sec:4_metrics_and_benchmark}

Evaluating visual reinforcement learning (RL) with large models requires both traditional RL metrics~\cite{jordan2020evaluating,agarwal2021deep} and new ones designed for complex, open-ended outputs.
Metrics like cumulative return and task success rate are still useful especially in tasks involving agents and environments but they are increasingly combined with preference-based evaluations.
In this section, we discuss metrics and benchmarks across four major application areas of large-model RL.

\subsection{Formalizing Metric Granularity}
\label{sec:metric_formal}

\noindent
Let $\mathcal{P}_{\text{test}}=\{\,p_1,\dots,p_D\}$ denote a fixed set of
\emph{prompts} (inputs) and let a generative policy
$\pi_{\theta}(y\mid p)$ produce an \emph{output} $y$ (text, image, video, \emph{etc.})
conditioned on each prompt $p\in\mathcal{P}_{\text{test}}$.
As shown in Figuer~\ref{fig:metric}, we distinguish three granularities of evaluation:

\textbf{Task Metric \(\Mset\) (Set-level).}
As illustrated in Fig.~\ref{fig:metric} (a), set-level metrics evaluate the generative policy $\pi_\theta$ over the full prompt set $\mathcal{P}_{\text{test}}$, by comparing the distribution of generated outputs $\mathcal{Y}_{\text{gen}} = \{\pi_\theta(\cdot \mid p_i)\}_{i=1}^D$ to a reference set of ground-truth outputs $\mathcal{Y}_{\text{gt}} = \{y^{i}_{\text{gt}}\}_{i=1}^D$.
When the evaluation function operates \textit{without access to ground-truth outputs} such as CLIPScore~\cite{hessel2021clipscore}, we define:
\begin{equation}
\begin{split}
\Mset
  = \frac{1}{D} \sum_{i=1}^{D}
    \mathbb{E}_{y \sim \pi_\theta(\cdot \mid p_i)}
    \left[ \Msamp(y^{i}_{\text{gen}}, p_i) \right],
\end{split}
\label{eq:set_metric_noref}
\end{equation}
where \(\Msamp(y, p)\) denotes a sample-level reward function applied to each generated output.
By contrast, many classical vision metrics do rely on reference outputs such as IoU~\cite{rezatofighi2019generalized}.
For these cases, set-level evaluation is defined as:
\begin{equation}
\Mset= \frac{1}{D} \sum_{i=1}^{D}
    \mathbb{E}_{y \sim \pi_\theta(\cdot \mid p_i)} \left[ \Msamp(y^{i}_{\text{gen}}, y^{i}_{\text{gt}}, p_i) \right],
\label{eq:set_metric_ref}
\end{equation}
where each ground-truth output \(y^{i}_{\text{gt}}\) denotes $i$-th ground truth from the reference set.

\textbf{Reward/Advantage metric $\Msamp$ (Sample-level).}
As illustrated in Fig.~\ref{fig:metric} (b), reward and advantage metrics \(\Msamp\) operate at the granularity of individual input-output pairs, forming the backbone of reinforcement learning in generative settings.
Given a prompt \(p_i\), the policy \(\pi_\theta\) generates a sample \(y^i_{\text{gen}}\), which is then scored by \(\Msamp(y^i_{\text{gen}}, p_i)\) to compute a scalar reward or advantage signal.
This feedback is used to optimize the policy via reinforcement learning (\eg~PPO~\cite{schulman2017proximal}, DPO~\cite{rafailov2023direct}).
In preference-based learning, the sample-level metric is often learned from human or GPT-4 comparisons~\cite{ouyang2022training, sun2023aligning}, or automatically derived via scoring models like CLIPScore~\cite{hessel2021clipscore}, or ImageReward~\cite{xu2023imagereward}.
When rewards are reference-dependent (e.g., using PSNR~\cite{wang2004image} or IoU~\cite{rezatofighi2019generalized}), \(\Msamp\) compares the generated output \(y^i_{\text{gen}}\) to a ground-truth output \(y^i_{\text{gt}}\).
Formally, the reward signal can be expressed as:
\begin{equation}
\Msamp(y^i_{\text{gen}}, p_i) = R_\phi(y^i_{\text{gen}}, p_i) 
\quad \text{or} \quad 
R_\phi(y^i_{\text{gen}}, y^i_{\text{gt}}, p_i),
\end{equation}
depending on whether the reward model \(R_\phi\) operates with or without access to ground-truth outputs.
In practice, the per-sample scores are transformed into step-wise advantages
\(\hat{A}_{i,t}\) (where \(t\) indexes generation steps).
These advantages directly drive policy updates, enabling reward shaping
and exploration control at the level of individual outputs.

\textbf{State-level Metric \(\Mstate^{t}\).}
As depicted in Fig.~\ref{fig:metric} (c), state-level metrics monitor
the \emph{training dynamics} of the current policy
\(\pi_{\theta_t}\) at iteration \(t\).
A common choice is the KL divergence to a frozen reference policy
\(\pi_{\theta_0}\):
\begin{equation}
\Mstate^{t} =
\mathbb{E}_{p\sim\mathcal{P}_{\text{val}}}
    D\!\bigl(
      \pi_{\theta_t}(\cdot\mid p)\,
      \Vert\,
      \pi_{\theta_0}(\cdot\mid p)
    \bigr).
\label{eq:state_metric_kl}
\end{equation}
Other diagnostics include \emph{output-length drift} for
autoregressive models and \emph{DDIM step-trace variance}
for diffusion models.  By tracking \(\Mstate^{t}\) during optimisation,
practitioners detect reward hacking, mode collapse, or excessive policy
shift before these issues degrade final performance.

\subsection{Evaluation of Multi-Modal Large Language Models}
\label{sec:eval_mllms}

\textbf{Task Metric.}  
As summarized in Table~\ref{tab:metrics_visual_rl}, MLLM are first judged on \emph{external, RL-free benchmarks}. 
General reasoning suites such as \textsc{MME}~\cite{fu2023mme}, \textsc{SEED-Bench}~\cite{li2024seed} and \textsc{MMBench}~\cite{liu2024mmbench} measure factual QA, commonsense and multi-step chain-of-thought across images.
Domain-specific subsets probe OCR (TextVQA~\cite{singh2019towards}), mathematics (MathVista~\cite{lu2023mathvista}), documents (ChartQA~\cite{lu2023mathvista}) and multilingual grounding (CMMMU~\cite{zhang2024cmmmu}).
%

\textbf{Reward Metric.}  
During training, each generated answer is scored with a sample-level reward \(\Msamp\).  
Three sources dominate current practice.
(i)~\emph{Human-preference rewards} are learned from large RLHF corpora ~\eg InstructGPT~\cite{ouyang2022training} and LLaVA-RLHF~\cite{sun2023aligning}, and give dense feedback that closely matches user judgements. 
(ii)~\emph{Verifiable rewards} arise when a sub-task admits deterministic checks, such as unit-test pass rate in CodeRL~\cite{le2022coderl} or symbolic exactness in DeepSeekMath~\cite{shao2024deepseekmath}; they are noise-free but limited in scope.
(iii)~\emph{Model-preference rewards} replace humans with a stronger frozen critic, \eg  CriticGPT~\cite{mcaleese2024llm}, delivering scalable but potentially biased supervision.
The chosen reward is converted to advantages \(\hat{A}_{i,t}\) and optimized via PPO, GRPO or DPO.

\textbf{Model-State Metric.}  
Beyond external scores, practitioners track light-weight diagnostics \(\Mstate^{t}\) throughout RL updates.  
Two lightweight diagnostics are widely adopted:
(i)~\emph{Output length drift}, the deviation of answer length from the SFT
baseline large drift foreshadows verbosity or repetition~\cite{rafailov2023direct};
(ii)~\emph{KL divergence} between the current policy $\pi_{\theta_t}$ and
frozen SFT reference $\pi_{\theta_0}$, as used in
InstructGPT~\cite{ouyang2022training}.  

\subsection{Evaluation of Visual Generation Models}
\label{sec:eval_visual_gen}

\textbf{Task Metric.}  
As listed in Table~\ref{tab:metrics_visual_rl}, final quality is judged on standard, RL–free benchmarks that target complementary axes.
Image fidelity \& diversity is measured by FID and Inception Score, while pixel-level reconstruction tasks (super-resolution, inpainting) use PSNR or SSIM.  
For prompt alignment, CLIP Score and Fréchet CLIP Distance quantify semantic correctness; video models additionally report FVD or Video IS to capture temporal coherence.  

\textbf{Reward Metric.}  
During RL fine-tuning, each generated image or video receives a sample-level reward \(\Msamp\).  
Human-preference rewards, \eg ImageReward \cite{xu2023imagereward} and HPS \cite{wu2023human} supply dense signals that correlate well with aesthetic appeal.
When a deterministic checker exists, authors turn to \emph{verifiable rewards}: MotionPrompt~\cite{nam2025optical} and DSPO \cite{cai2025dspo} use optical-flow, object masks that can be evaluated without humans.  
A third route relies on model preference rewards, where a stronger frozen critic (\eg VideoPrefer \cite{wu2024boosting} or PARM \cite{zhang2025let}) scores samples, enabling scalable DPO/PPO training.

\textbf{Model-State Metric.}  
Two light diagnostics track training stability.  
(i)~Denoising trajectory statistics: VARD \cite{dai2025vard} and Inversion DPO \cite{li2025inversion} record per-step noise predictions or DDIM traces; pathological spikes reveal early collapse.  
(ii)~KL divergence between the current diffusion policy and its frozen base (\(\pi_{\theta_t}\!\parallel\!\pi_{\theta_0}\)), popularized by DDPO \cite{black2023ddpo} and reused in VARD.  
%

\subsection{Evaluation of Unified Models}
\label{sec:eval_unified}

\textbf{Task Metric.}  
Two benchmark families are widely adopted.
\emph{Generation-oriented suites} such as
GenEval~\cite{ghosh2023geneval},
DPG-Bench~\cite{hu2024ella},
and ParaPrompts~\cite{wu2025paragraph}
focus on prompt-faithful generation, testing multi-object composition,
style control, and long-caption adherence.
Conversely, understanding-oriented benchmarks
(\textsc{MME}~\cite{fu2023mme}, \textsc{POPE}~\cite{li2023evaluating})
measure grounding, reasoning and hallucination detection from the same
backbone.
%

\textbf{Reward Metric.}  
Recent work explores two design philosophies for training signals.
\emph{Unified rewards} (~\eg~  UniRL~\cite{mao2025unirl},
CoRL~\cite{jiang2025co}) blend multiple objectives textual correctness,
CLIP alignment, aesthetic quality—into a single scalar that drives one
shared policy across tasks.
In contrast, task-specific rewards keep the generator and
understanding heads separate, applying RL only to the generation branch as
in \textsc{Vargpt-v1.1}~\cite{zhuang2025vargpt} or Emu3~\cite{wang2024emu3}.
The former promotes cross-modal transfer, while the latter preserves the
stability of perception modules.

\textbf{Model-State Metric.}  
Unified models additionally track fine-grained diagnostics during RL.
UniRL~\cite{mao2025unirl} proposes a
generation–understanding imbalance score the absolute gap between
batch-level rewards on the two task families to prevent one modality from
dominating the update.
HermesFlow~\cite{yang2025hermesflow} monitors the
KL divergence between the current shared policy
\(\pi_{\theta_t}\) and its supervised baseline
\(\pi_{\theta_0}\) on \emph{both} generation and understanding prompts,
serving as an early-warning signal for policy collapse.
These state-level curves \(\Mstate^{t}\) allow practitioners to
stop or re-weight training before external task scores degrade.

\subsection{Evaluation of Vision Language Action Models}

\textbf{Task Metric.}
In GUI Automation task, there are multiple benchmarks could be classified into online or offline scenarios.
For \textit{offline} setting, it mainly have grounding and navigation parts. For grounding, mainly check whether the click action fail into the target button; For navigation, it requires model to predict current action conditioned on oracle past history, this mainly dependent on whether the action class (click or type) are correctly predicted per step.
For \textit{online} setting, it is more challenging, because it requires the model to fully perform multi-step execution which is long procedural then check whether the final outcome meet the task requirement. Such long procedural setups will produce sparse signal in term of model evalation.

\textbf{Reward Metric.}
For reward modeling, most offline RL methods borrow the metric from task metric like IoU, while come to the online environment, due to the sparsity of task success rate, which present significant challenges for end-to-end multi-turn RL training~\cite{arpo} i.e., lack of training efficency, lack of informativeness, step-level reward is proposed to address this issue, such as developing a reward or critic models~\cite{digirl,webworld}.

\textbf{Model-State Metric.}  
To fully understand the model behavior beyond task success rate, trajectory length being an important metric as it can reflect how efficient model can address the task. A smart agent should be able to resolve the task with minimal steps. This pose challenges for agents with advanced planning ability.

\begin{table*}[t]
    \centering
      \scriptsize
      \caption{\textbf{Public benchmarks of MLLM most commonly used in visual RL.}  
    Only benchmarks relevant to visual reinforcement learning are included (RL-focused training \& evaluation); task-specific benchmarks, such as MME~\cite{fu2023mme}, are excluded from consideration.
    `Tr' and `Te' refer to the `Train' and `Test', respectively. 
    }
    \label{tab:benchmarks_vrl}
    \input{table_bench}
\end{table*}

\begin{table*}[t]
    \centering
      \scriptsize
      \caption{\textbf{Public benchmarks of Visual Generation
(image/video/3D) most commonly used in visual RL.}  
    Only benchmarks relevant to visual reinforcement learning are included (RL-focused training \& evaluation).
    `Tr' and `Te' refer to the `Train' and `Test', respectively. 
    }
    \label{tab:benchmarks_vrl_visualgen}
    \input{table_bench_2}
\end{table*}

\begin{table*}[t]
    \centering
      \scriptsize
      \caption{\textbf{Public benchmarks of VLAs most commonly used in visual RL.}  
    Only benchmarks relevant to visual reinforcement learning are included (RL-focused training \& evaluation).
    `Tr' and `Te' refer to the `Train' and `Test', respectively. 
    }
    \label{tab:benchmarks_vla}
    \input{table_bench_3}
\end{table*}

\subsection{Benchmarks}

A variety of new benchmarks explicitly support RL-based training and evaluation in the visual domain (see Table~\ref{tab:benchmarks_vrl}).
For \textbf{MLLM}, recent datasets target complex multi-hop reasoning and alignment with human preferences.
For example, SEED-Bench-R1~\cite{chen2025exploring} introduces a hierarchical egocentric video question-answering benchmark with $50$k training questions and a human-verified validation set.
Long Video RL~\cite{chen2025scaling} scales up multi-step reasoning on long videos: it provides $52$k QA pairs with detailed reasoning annotations.
Another recent benchmark, Ego-R1 Bench~\cite{tian2025egor1chainoftoolthoughtultralongegocentric}, focuses on ultra-long (week-long) egocentric videos; an RL-trained “chain-of-tool-thought” agent must invoke perception tools in ~$7$ sequential steps on average to answer each query, illustrating the use of step-wise reasoning accuracy as a core challenge.
In the image domain, VisuLogic~\cite{xu2025visulogic} contains $1,000$ carefully crafted visual reasoning puzzles (\eg~ spatial and logic problems) to evaluate pure vision-centric reasoning, and most models perform only slightly above random on this benchmark.

Benchmarks for \textbf{visual generation} tasks predominantly supply human preference data that serve as reward models for policy optimization. 
Datasets like ImageReward~\cite{xu2023imagereward} and HPS v1 \& v2~\cite{wu2023human,wu2023human} collect human-ranked pairs of text-to-image outputs, allowing one to train a scalar reward function that scores generations.
Such reward models have been used to refine text-to-image diffusion models via RLHF, aligning outputs with human aesthetic preferences.
Similarly, Pick-a-Pic~\cite{kirstain2023pick} and VideoReward ~\cite{liu2025improving}extend this to broader user preferences (motion smoothness, text alignment).
Some benchmarks also facilitate robust evaluation of generative RL agents on generalization.
T2I-CompBench~\cite{huang2023t2i} is a text-to-image compositionality test set that requires correctly binding novel combinations of attributes and object relations – a measure of compositional generalization often used to assess RL-trained generators.
Likewise, domain-specific benchmarks define verifiable success criteria as rewards: StarVector~\cite{rodriguez2025starvector} provides SVG code-generation tasks with a strict shape-matching reward, and AnimeReward~\cite{zhu2025aligning} targets consistency in animated video generations with multi-dimensional human preference scores (image-video coherence, character consistency, etc.).

For \textbf{vision–language action} agents, numerous benchmarks provide expert trajectories and simulated environments with clear reward signals for policy training and robust evaluation.
Many are centered on GUI and web interaction tasks, where success can be unambiguously measured.
For instance, GUI-R1-3K~\cite{luo2025gui} compiles $3,000+$ GUI manipulation trajectories across Windows, Linux, macOS, Android, and web platforms.
It introduces an “R1-style” dense reward scheme mapping each correct action (\eg clicking the right button, entering correct text, a valid scroll) to positive feedback, providing step-wise reinforcement that guides an agent through multi-step UI tasks.
Building on this, SE-GUI~\cite{yuan2025enhancing} curates 3k high-quality GUI examples with grounded instructions and bounding-box annotations, which are used to train agents with a self-imitation RL strategy.
Evaluation-focused suites like UI-R1~\cite{lu2025ui} define a fixed set of unseen tasks (\eg~ 136 mobile GUI tasks spanning click, scroll, swipe, text-input actions) to test generalization of learned policies.
Meanwhile, web interaction benchmarks such as Mind2Web~\cite{deng2023mind2web} offer ~2,000 tasks on real websites with a binary success/failure reward for completing each task.
Some datasets emphasize exact match and reproducibility: AITZ~\cite{zhang2024android} (Android Interaction w/ CoAT reasoning) logs ~18k screen-action pairs with corresponding tool-assisted rationales, and uses an exact action match reward to ensure precise adherence to instructions.
On the other hand, broader benchmarks like OmniAct~\cite{kapoor2024omniact} and GUICoURS~\cite{chen2024guicourse} target generalist agent capabilities across diverse domains.
OmniAct integrates nearly 10k scripted desktop and web tasks into a single environment, while GUICoURS combines multimodal resources (10M OCR observations, 67k navigation demonstrations) spanning GUI, web, and chat interfaces.
%
%
The reward structures in these benchmarks are carefully crafted, from rule-based metrics to preference scores, to guide policy learning and reflect task goals.
They enable visual RL agents to learn from meaningful feedback and be evaluated not just on task success, but also on alignment with human reasoning and performance on complex, long-horizon decisions.

%% file: table_metric.tex
\renewcommand{\arraystretch}{1.3}
\begin{tabularx}{\textwidth}{>{\raggedright\arraybackslash}p{2cm} >{\raggedright\arraybackslash}X >{\raggedright\arraybackslash}X >{\raggedright\arraybackslash}X}
\toprule
\textbf{Task Family} & \textbf{Task Metric} $\Mset$ \newline (RL-free benchmarks / scores) & \textbf{Reward / Advantage Estimation} $\Msamp$ \newline (scoring \& preference signals) & \textbf{Model-state Metric} $\Mstate^{t}$  \newline (training diagnostics)  \\
\midrule

\textbf{Multimodal LLMs/ VLMs} &
• \textbf{Comprehensive Evaluation}: \eg~MME~\cite{fu2023mme}, SEED-Bench~\cite{li2024seed}, VQA v2~\cite{goyal2017making}, MMBench~\cite{liu2024mmbench}. 
\newline
•  \textbf{OCR}: \eg~TextVQA~\cite{singh2019towards}, OCR-VQA~\cite{mishra2019ocr}, OCRBench~\cite{liu2023hidden}.
\newline
•  \textbf{Mathematical}: \eg~MathVista~\cite{lu2023mathvista}.
\newline
•  \textbf{Documentation}: \eg~ChartQA~\cite{lu2023mathvista}, DocVQA~\cite{mathew2021docvqa}, InfoVQA~\cite{mathew2022infographicvqa}.
\newline
•  \textbf{Multilingual}: \eg~CMMMU~\cite{zhang2024cmmmu}, CMMU~\cite{he2024cmmu}.
\newline
$\dotsc$
&
• \textbf{Reward from Human Preference}: \eg~InstructGPT~\cite{ouyang2022training}, LLaVA-RLHF~\cite{sun2023aligning}.
\newline
• \textbf{Verifiable Rewards}: Deepseekmath~\cite{shao2024deepseekmath}, IoU~\cite{rezatofighi2019generalized}, CodeRL~\cite{le2022coderl}.
\newline
• \textbf{Reward from Model Preference}:~\eg CriticGPT~\cite{mcaleese2024llm}.
\newline
$\dotsc$
&
• \textbf{Output length monitoring}: \eg~DPO~\cite{rafailov2023direct}.
\newline
• \textbf{KL-divergence}: \eg~InstructGPT~\cite{ouyang2022training}.
\newline
$\dotsc$
\\

\midrule

\textbf{Visual Generation} \newline (Image / Video / 3D) 
&
• \textbf{Image Fidelity \& Diversity}: \eg~FID~\cite{heusel2017gans}, Inception Score (IS)~\cite{salimans2016improved}. \newline
• \textbf{Pixel-level Reconstruction}: \eg~PSNR~\cite{wang2004image}, SSIM~\cite{wang2004image}.
\newline
• \textbf{Semantic Alignment}: \eg~Geneval~\cite{ghosh2023geneval}, T2I-CompBench~\cite{huang2023t2i}, CLIP Score~\cite{radford2021learning}, Frechet CLIP Distance~\cite{betzalel2022study}.
\newline
• \textbf{Video Fidelity \& Diversity}: \eg~FVD~\cite{unterthiner2019fvd}, Video IS~\cite{saito2020train}.
\newline
$\dotsc$
&
• \textbf{Reward from Human Preference}:~\eg ImageReward~\cite{xu2023imagereward}, HPS~\cite{wu2023human}, HPS V2~\cite{wu2023human}, Pick-a-pic~\cite{kirstain2023pick}, VideoReward~\cite{liu2025improving}, RichHF-18K~\cite{liang2024rich}.
\newline
• \textbf{Verifiable Rewards}:~\eg MotionPrompt~\cite{nam2025optical}, DSPO~\cite{cai2025dspo}, Instructrl4pix~\cite{li2024instructrl4pix}.
\newline
• \textbf{Reward from Model Preference}:~\eg VideoPrefer~\cite{wu2024boosting}, PARM~\cite{zhang2025let}.
\newline
$\dotsc$
&
• \textbf{Denoising Trajectory Diagnostics}: \eg~VARD~\cite{dai2025vard}, Inversion-DPO~\cite{li2025inversion}.
\newline
• \textbf{KL-divergence}: \eg~DDPO~\cite{black2023ddpo}, VARD~\cite{dai2025vard}.
\newline
$\dotsc$

\\

\midrule

\textbf{Unified Models} &
• \textbf{Generation–oriented Task Metrics}:~\eg GenEval~\cite{ghosh2023geneval}, DPG-Bench~\cite{hu2024ella}, ParaPrompts~\cite{wu2025paragraph}. 
\newline
• \textbf{Understanding–oriented Task Metrics}:~\eg MME~\cite{fu2023mme}, POPE~\cite{li2023evaluating}.
\newline
$\dotsc$
&
• \textbf{Unified Rewards}:~\eg UniRL~\cite{mao2025unirl}, CoRL~\cite{jiang2025co}. \newline
• \textbf{Task-specific Rewards}:~\eg Vargpt-v1. 1~\cite{zhuang2025vargpt}, Emu3~\cite{wang2024emu3}. \newline
$\dotsc$
 &
• \textbf{Generation–Understanding Imbalance}:~\eg UniRL~\cite{mao2025unirl}. \newline
• \textbf{KL-divergence}:~\eg HermesFlow~\cite{yang2025hermesflow}
\newline
$\dotsc$
\\

\midrule

\textbf{Vision-Language Action Agents} \newline (GUI / Navigation / Manip.) &
• \textbf{GUI Action Accuracy (offline)}:~\eg ScreenSpot~\cite{seeclick}, ScreenSpot-Pro~\cite{screenspotpro}, Ui-vision~\cite{uivision}.
\newline
• \textbf{Task Success rate (online):}~\eg Webarena~\cite{webarena}, Osworld~\cite{osworld}, Windows agent arena~\cite{windowarena}.
\newline
• \textbf{Model-based Evaluation:}~\eg Agentrewardbench~\cite{ agentrewardbench}, Webworld~\cite{webworld},  Digirl~\cite{digirl}. 
\newline
• \textbf{Reward from Human Preference}:~\eg
RFTF~\cite{shu2025rftf}.
\newline
• \textbf{Navigation Success \& SPL:}
\newline
$\dotsc$
&
• \textbf{Rule-based Success (binary) (IoU, Action Accuracy):}~\eg UI-R1~\cite{lu2025ui}, ARPO~\cite{arpo}, VLA-RL~\cite{lu2025vla}.
\newline
• \textbf{Model Preference Critic:}~\eg ProgRM~\cite{zhang2025progrm}.
\newline
• \textbf{Dense Shaping (distance or coverage):}~\eg LPO~\cite{tang2025lpo}, Gui-r1~\cite{luo2025gui}.
\newline
$\dotsc$
&
• \textbf{Action trajectory length:}~\eg Osworld~\cite{osworld}.
\newline
• \textbf{KL penalty for policy stability}:~\eg GUI-R1~\cite{luo2025gui}, UI-R1~\cite{lu2025ui}. 
\newline
• \textbf{Output length monitoring:}~\eg UI-R1~\cite{lu2025ui}.
$\dotsc$\\

\bottomrule
\end{tabularx}

%% file: table_bench.tex
\begin{tabularx}{\textwidth}{>{\raggedright\arraybackslash}p{2cm}
                                 >{\raggedright\arraybackslash}p{2.4cm}
                                 >{\raggedright\arraybackslash}p{1.2cm}
                                 >{\raggedright\arraybackslash}p{0.5cm}
                                 >{\raggedright\arraybackslash}p{10.3cm}}
\toprule
\textbf{Task family} & \textbf{Benchmark} & \textbf{Date} & \textbf{Tr/Te} & \textbf{Description} (benchmark info. and RL reward signal)  \\
\midrule
\multirow{28}{=}{\textbf{Multimodal} \\ \textbf{LLMs / VLMs}} &
\textbf{SEED-Bench-R1}~\cite{chen2025exploring} & Mar 2025 & Tr\&Te &
Video-QA pairs with human-preference reward model (used in~\cite{chen2025exploring})
\\
& \textbf{Video-Holmes}~\cite{cheng2025video}  & May 2025 & Te  & Human-ranked T2I pairs from diverse generation models (used in~\cite{cheng2025video})\\
& \textbf{VisuLogic}~\cite{xu2025visulogic}  & Apr 2025 & Tr\&Te  &  Visual-reasoning QA set; exact-match reward enables RL fine-tuning (used in~\cite{xu2025visulogic})\\
& \textbf{R1-ShareVL}~\cite{yao2025r1sharevl}  & May 2025 & Tr  &  52 k MM-Eureka subset for Share-GRPO RL training (used in~\cite{yao2025r1sharevl})\\
& \textbf{Ego-R1}~\cite{tian2025egor1chainoftoolthoughtultralongegocentric}  & Jun 2025 & Tr\&Te  &  25 k CoTT egocentric traces enabling RL training for ultra-long video reasoning (used in~\cite{tian2025egor1chainoftoolthoughtultralongegocentric})\\
& \textbf{Long-RL}~\cite{chen2025scaling}  & Jul 2025 & Tr\&Te  &  104 K long-video QA pairs (GRPO accuracy / format reward) (used in~\cite{chen2025scaling})\\
& \textbf{VisCOT}~\cite{shao2024visual}  & Mar 2024 & Tr\&Te  &  438 k visual chain-of-thought traces with step-wise rewards for RL (used in~\cite{shao2024visual})\\
& \textbf{MMK12}~\cite{meng2025mm}  & Mar 2025 & Tr\&Te  &  15.6 K multimodal math problems (rule-based accuracy / format rewards)  (used in~\cite{meng2025mm})\\
& \textbf{Time-R1}~\cite{wang2025timer1}  & Mar 2025 & Tr\&Te  &  2.5 K TimeRFT grounding spans (IoU reward)  (used in~\cite{wang2025timer1})\\
& \textbf{VSI-Bench}~\cite{yang2024think}  & Dec 2024 & Te  &  Spatial QA benchmark offering RL exact-match reward  (used in~\cite{yang2024think})\\
& \textbf{MME-Reasoning}~\cite{yuan2025mme}  & May 2025 & Te  &  Logic QA benchmark  (used in~\cite{yuan2025mme})\\
& \textbf{K12-2M}~\cite{wang2025mathcodervl}  & May 2025 & Tr  &  2 M multimodal math pairs for RL training (used in~\cite{wang2025mathcodervl})\\
& \textbf{MathVista}~\cite{lu2024mathvista}  & Oct 2023  & Te  &  Visual-math QA benchmark; test-only RL exact-match reward (used in~\cite{lu2024mathvista})\\
& \textbf{MathVerse}~\cite{zhang2024mathverse}  & Mar 2024  & Te  &  Diagram-math QA benchmark; test-only RL exact-match reward (used in~\cite{zhang2024mathverse})\\
& \textbf{EMMA}~\cite{hao2025can}  &  Jan 2025  & Te  &  Robust multimodal reasoning benchmark challenging current MLLMs (used in~\cite{hao2025can})\\
& \textbf{WeMath}~\cite{qiao2024we}  &  Jul 2024  & Te  &  Multimodal math reasoning benchmark with diagrams (used in~\cite{qiao2024we})\\
& \textbf{DynaMATH}~\cite{zou2024dynamic}  &  Oct 2024  & Te  &  Dynamic visual math reasoning robustness benchmark for VLMs (used in~\cite{zou2024dynamic})\\
& \textbf{MM-IQ}~\cite{cai2025mm}  &   Feb 2025  & Te  &  Human-like multimodal abstraction and reasoning benchmark (used in~\cite{cai2025mm})\\
& \textbf{OlympiadBench}~\cite{he2024olympiadbench}  &   Feb 2024  & Te  &  Bilingual multimodal Olympiad-level scientific reasoning benchmark (used in~\cite{he2024olympiadbench})\\
& \textbf{ZeroBench}~\cite{roberts2025zerobench}  &   Feb 2025  & Te  &  near-impossible visual reasoning stress test for LMMs (used in~\cite{roberts2025zerobench})\\
& \textbf{MMMU-Pro}~\cite{yue2024mmmu}  &   Sep 2024  & Te  &  robust multidisciplinary multimodal understanding benchmark (used in~\cite{yue2024mmmu})\\
& \textbf{MME-CoT}~\cite{jiang2025mme}  &   Feb 2025  & Te  &  Multimodal chain-of-thought reasoning benchmark dataset (used in~\cite{jiang2025mme})\\
& \textbf{MMIR}~\cite{yan2025multimodalinconsistencyreasoningmmir}  &  Feb 2025  & Te  &  Multimodal inconsistency reasoning benchmark dataset (used in~\cite{yan2025multimodalinconsistencyreasoningmmir})\\
& \textbf{SpatialEval}~\cite{wang2024spatial}  &  Jun 2024  & Te  &  Synthetic spatial reasoning benchmark for VLMs (used in~\cite{wang2024spatial})\\
& \textbf{MMReason}~\cite{wang2024spatial}  &  Jun 2025   & Te  &  Synthetic spatial reasoning benchmark for VLMs (used in~\cite{wang2024spatial})\\
& \textbf{IntentBench}~\cite{yang2025humanomniv2}  &   Jun 2025   & Te  &  Omnimodal evaluation demands unified audio-visual comprehension (used in~\cite{yang2025humanomniv2})\\

\midrule
\multirow{10}{=}{\textbf{Visual Generation}\\(image/video/3D)} &
\textbf{ImageReward}~\cite{xu2023imagereward} & Apr 2023 & Tr\&Te &  Human-ranked pairs for T2I; reward model outputs scalar score (used in~\cite{wu2025reprompt,xiao2024diffppo,zhou2025multimodalllmscustomizedreward}) \\
& \textbf{HPS}~\cite{wu2023human}  & Mar 2023 & Tr\&Te  & Human-ranked T2I pairs from diverse generation models (used in~\cite{wu2023human})\\
& \textbf{HPS V2}~\cite{wu2023humanv2}  & Jun 2023 & Tr\&Te & Human-ranked T2I pairs across diverse prompts, preference-based reward (used in~\cite{duan2025got,gu2024diffusion})\\
& \textbf{Pick-a-Pic}~\cite{kirstain2023pick}  & May 2023 & Tr\&Te & User preferences for pairwise rankings over T2I generations (used in~\cite{wallace2024diffusion,li2024aligning,gu2024diffusion,lee2024parrot}) \\
& \textbf{VideoReward}~\cite{liu2025improving}  & Jan 2025 & Te & Human-ranked video pairs across quality, motion, and text alignment (used in~\cite{liu2025improving})\\
& \textbf{T2I-CompBench}~\cite{huang2023t2i}  & Jul 2023 & Tr\&Te & Compositional text-to-image dataset covering attributes and object relations. (used in~\cite{gupta2025simple,duan2025got}) 
\\
& \textbf{StarVector}~\cite{rodriguez2025starvector}  & Dec 2023 & Tr\&Te & SVG Code Generation Data, Match Reward (used in~\cite{rodriguez2025rendering})
\\
& \textbf{AnimeReward}~\cite{zhu2025aligning}  & Apr 2025 & Tr & Multi‑dimensional (\eg character consisten) human preference anime videos (used in~\cite{zhu2025aligning})
\\
& \textbf{VideoPrefer}~\cite{wu2024boosting}  & Dec 2024 & Tr & MLLM‑annotated 135K video preference pairs (used in~\cite{wu2024boosting})
 \\
\midrule


\multirow{14}{=}{\textbf{Vision–Language \\ Action Agents}} &
\textbf{GUI-R1-3K} \cite{luo2025gui}  & Apr 2025 & Tr\&Te & GUI trajectories spanning Windows, Linux, macOS, Android, and Web platforms  (used in~\cite{luo2025gui}) \\

&\textbf{SE-GUI-3k} \cite{yuan2025enhancing}  & May 2025 & Tr & 	3,018 examples (desktop / web / mobile) with instruction and bounding box (used in~\cite{yuan2025enhancing})  \\

&\textbf{UI-R1}~\cite{lu2025ui}  & May 2025 & Tr &
136 mobile GUI tasks covering 5 action types (click, scroll, swipe, text-input) (used in~\cite{yuan2025enhancing})\\

&\textbf{CAGUI}~\cite{zhang2025agentcpm} & Jun 2025 & Te &
55 K trajectories from 30 Chinese Android apps, 8 domains (used in~\cite{yuan2025enhancing})\\

&\textbf{Mobile-R1}~\cite{gu2025mobile} & Jun 2025 & Tr\&Te &
More than 500 online task trajectories from 28 Chinese apps (used in~\cite{gu2025mobile})\\

&\textbf{Mind2web}~\cite{deng2023mind2web} & Jun 2023 & Tr\&Te & 2 k tasks on 137 real websites; success/fail reward for RL (used in~\cite{deng2023mind2web,tang2025lpo})\\

&\textbf{AITZ}~\cite{zhang2024android} & Jun 2023 & Tr\&Te & 18,643 Android screen–action pairs with CoAT reasoning (used in~\cite{zhang2024android})\\

&\textbf{Omniac}~\cite{kapoor2024omniact} & Feb 2024 & Tr\&Te & Desktop + web ~9.8 k scripted tasks (used in~\cite{kapoor2024omniact})\\
 
&\textbf{GUICours}~\cite{chen2024guicourse} & Jun 2024 & Tr\&Te & GUIEnv/Act/Chat datasets (10 M OCR + 67 k navigation) (used in~\cite{chen2024guicourse})\\

&\textbf{Habitat 3.0}~\cite{puig2023habitat} & Oct 2023  & Tr\&Te & Interactive embodied-AI scenes with humans and robots (used in~\cite{puig2023habitat})\\

&\textbf{VLN-CE}~\cite{krantz2020beyond} & Apr 2020  & Tr\&Te & Continuous embodied navigation dataset with language instructions (used in~\cite{krantz2020beyond})\\

&\textbf{Rlbench}~\cite{james2020rlbench} & Apr 2020  & Tr\&Te & 
Multi-task simulated robot manipulation benchmark dataset (used in~\cite{james2020rlbench})\\

&\textbf{RoboCasa}~\cite{nasiriany2024robocasa} & Jun 2024   & Tr & Large-scale kitchen-task simulation for generalist robotics (used in~\cite{nasiriany2024robocasa})\\

&\textbf{LIBERO}~\cite{liu2023libero} & Jun 2023   & Tr & Lifelong robot learning benchmark with 100 manipulation tasks (used in~\cite{liu2023libero})\\

&\textbf{VLABench}~\cite{zhang2024vlabench} & Dec 2024   & Tr\&Te & Long-horizon language-conditioned manipulation benchmark for robots (used in~\cite{zhang2024vlabench})\\

\bottomrule
\end{tabularx}

%% file: table_bench_2.tex
\begin{tabularx}{\textwidth}{>{\raggedright\arraybackslash}p{5cm}
                                 >{\raggedright\arraybackslash}p{1.2cm}
                                 >{\raggedright\arraybackslash}p{0.7cm}
                                 >{\raggedright\arraybackslash}p{8.3cm}}
\toprule
 \textbf{Benchmark} & \textbf{Date} & \textbf{Tr/Te} & \textbf{Description} (benchmark info. and RL reward signal)  \\
\midrule

\textbf{ImageReward}~\cite{xu2023imagereward} & Apr 2023 & Tr\&Te &  Human-ranked pairs for T2I; reward model outputs scalar score (used in~\cite{wu2025reprompt,xiao2024diffppo,zhou2025multimodalllmscustomizedreward}) \\
\textbf{HPS}~\cite{wu2023human}  & Mar 2023 & Tr\&Te  & Human-ranked T2I pairs from diverse generation models (used in~\cite{wu2023human})\\
 \textbf{HPS V2}~\cite{wu2023humanv2}  & Jun 2023 & Tr\&Te & Human-ranked T2I pairs across diverse prompts, preference-based reward (used in~\cite{duan2025got,gu2024diffusion})\\
 \textbf{Pick-a-Pic}~\cite{kirstain2023pick}  & May 2023 & Tr\&Te & User preferences for pairwise rankings over T2I generations (used in~\cite{wallace2024diffusion,li2024aligning,gu2024diffusion,lee2024parrot}) \\
 \textbf{VideoReward}~\cite{liu2025improving}  & Jan 2025 & Te & Human-ranked video pairs across quality, motion, and text alignment (used in~\cite{liu2025improving})\\
 \textbf{T2I-CompBench}~\cite{huang2023t2i}  & Jul 2023 & Tr\&Te & Compositional text-to-image dataset covering attributes and object relations. (used in~\cite{gupta2025simple,duan2025got}) 
\\
\textbf{StarVector}~\cite{rodriguez2025starvector}  & Dec 2023 & Tr\&Te & SVG Code Generation Data, Match Reward (used in~\cite{rodriguez2025rendering})
\\ \textbf{AnimeReward}~\cite{zhu2025aligning}  & Apr 2025 & Tr & Multi‑dimensional (\eg character consisten) human preference anime videos (used in~\cite{zhu2025aligning})
\\ \textbf{VideoPrefer}~\cite{wu2024boosting}  & Dec 2024 & Tr & MLLM‑annotated 135K video preference pairs (used in~\cite{wu2024boosting})
 \\

\bottomrule
\end{tabularx}

%% file: table_bench_3.tex
\begin{tabularx}{\textwidth}{>{\raggedright\arraybackslash}p{4.5cm}
                                 >{\raggedright\arraybackslash}p{1.2cm}
                                 >{\raggedright\arraybackslash}p{0.7cm}
                                 >{\raggedright\arraybackslash}p{8.3cm}}
\toprule
 \textbf{Benchmark} & \textbf{Date} & \textbf{Tr/Te} & \textbf{Description} (benchmark info. and RL reward signal)  \\
\midrule

\textbf{GUI-R1-3K} \cite{luo2025gui}  & Apr 2025 & Tr\&Te & GUI trajectories spanning Windows, Linux, macOS, Android, and Web platforms   \\

\textbf{SE-GUI-3k} \cite{yuan2025enhancing}  & May 2025 & Tr & 	3,018 examples (desktop / web / mobile) with instruction and bounding box  \\

\textbf{UI-R1}~\cite{lu2025ui}  & May 2025 & Tr &
136 mobile GUI tasks covering 5 action types (click, scroll, swipe, text-input) \\

\textbf{CAGUI}~\cite{zhang2025agentcpm} & Jun 2025 & Te &
55 K trajectories from 30 Chinese Android apps, 8 domains \\

\textbf{Mobile-R1}~\cite{gu2025mobile} & Jun 2025 & Tr\&Te &
More than 500 online task trajectories from 28 Chinese apps \\

\textbf{Mind2web}~\cite{deng2023mind2web} & Jun 2023 & Tr\&Te & 2 k tasks on 137 real websites; success/fail reward for RL \\

\textbf{AITZ}~\cite{zhang2024android} & Jun 2023 & Tr\&Te & 18,643 Android screen–action pairs with CoAT reasoning \\

\textbf{Omniac}~\cite{kapoor2024omniact} & Feb 2024 & Tr\&Te & Desktop + web ~9.8 k scripted tasks \\
 
\textbf{GUICours}~\cite{chen2024guicourse} & Jun 2024 & Tr\&Te & GUIEnv/Act/Chat datasets (10 M OCR + 67 k navigation) \\

\textbf{Habitat}~\cite{puig2023habitat} & Oct 2023  & Tr\&Te & Interactive embodied-AI scenes with humans and robots \\

\textbf{VLN-CE}~\cite{krantz2020beyond} & Apr 2020  & Tr\&Te & Continuous embodied navigation dataset with language instructions \\

\textbf{RLBench}~\cite{james2020rlbench} & Apr 2020  & Tr\&Te & 
Multi-task simulated robot manipulation benchmark dataset \\

\textbf{RoboCasa}~\cite{nasiriany2024robocasa} & Jun 2024   & Tr & Large-scale kitchen-task simulation for generalist robotics \\

\textbf{LIBERO}~\cite{liu2023libero} & Jun 2023   & Tr & Lifelong robot learning benchmark with 100 manipulation tasks \\

\textbf{VLABench}~\cite{zhang2024vlabench} & Dec 2024   & Tr\&Te & Long-horizon language-conditioned manipulation benchmark for robots \\

\bottomrule
\end{tabularx}

%% file: 5_future.tex
\section{Challenges and Future Works}
\label{sec:5_future}

\subsection{Effective Reasoning: Balancing Depth and Efficiency}
A recurrent challenge in visual RL is reasoning calibration: excessively long chains of visual or verbal thoughts incur latency and compounding errors, whereas overly aggressive pruning discards salient cues.
We foresee two research thrusts.
(\textbf{i}) \emph{Adaptive horizon policies}: train a termination critic that jointly optimizes answer quality and computational cost; curriculum-based reward shaping can gradually penalize redundant steps while preserving informative ones.
(\textbf{ii}) \emph{Meta-reasoning and few-shot self-evaluation}: incorporate a lightweight evaluator that critiques partial chains (\eg via frozen vision–language models) and decides whether further thinking is worthwhile. 
Future benchmarks should therefore report both success rate and reasoning efficiency metrics (average steps, FLOPs, latency), encouraging algorithms that achieve high accuracy with just-enough deliberation rather than maximal cogitation.

\subsection{Long-Horizon RL in VLA}
Long-horizon vision–language agents (VLA) must execute dozens of atomic actions (\eg clicks, drags, text edits) before any end-task reward is observed.
Existing works such as OS-World~\cite{abhyankar2025osworld} and ARPO~\cite{arpo} therefore fall back on sparse reward for each click and a binary task success flag yet empirical results suggest that even GRPO yields limited gains under such supervision.
Future research should (i) \emph{discover intrinsic sub-goals}: segment trajectories via state-change detection or language-conditioned clustering, then assign dense rewards to sub-goal completions;
(ii) \emph{learn affordance critics}: train contrastive vision–language models to score how much an action reduces the distance to the verbal goal, providing shaped feedback without manual labels;
(iii) \emph{hierarchical or option-based RL}: couple a high-level language planner that proposes semantic sub-tasks with a low-level policy fine-tuned by off-policy RL or decision transformers; 
%

\subsection{RL for Thinking with Vision}

Recent works for visual planning, such as Chain-of-Focus~\cite{zhang2025chain} and Openthinkimg~\cite{su2025openthinkimg} all treat the picture as an external workspace: the agent may crop, sketch, highlight or insert visual tokens before emitting the next language token.
While early prototypes rely on supervised heuristics for these spatial actions, moving to reinforcement learning exposes four open problems. 
\textbf{(i) Action‐space design.}  Cropping or doodling is naturally continuous ($x,y,w,h,\dots$) yet RL libraries and GPU memories favor small discrete sets. 
Hybrid schemes that learn a differentiable proposal policy and then refine coordinates via policy gradient fine-tuning, as hinted by BRPO~\cite{chu2025qwen} and VRAG-RL~\cite{wang2025vrag}, remain largely unexplored.  
\textbf{(ii) Credit assignment.} 
Most benchmarks only reward the final task success (\eg answer correctness in VILASR~\cite{wu2025reinforcing}); the whole visual chain-of-thought shares a single sparse scalar.
Future work should mine step-wise proxy rewards, \eg CLIP similarity increase after a crop, or entropy drop in a learned belief state—to enable bootstrapped or hierarchical RL. 
\textbf{(iii) Data efficiency.}  Sketching or inserting patches triggers extra forward passes through the vision encoder, making naive on-policy RL prohibitively expensive.
Relabeling (DeepEyes~\cite{zheng2025deepeyes}) and model-based imagination (Pixel Reasoner~\cite{su2025pixel}) point to sample-efficient alternatives, but principled replay and uncertainty-aware planners for visual actions are still missing.  
Therefore, future directions include: learning structured visual skills (crop, zoom, draw) via skill-prior RL;
devising cross-modal reward shaping that scores each edit by how much it simplifies the remaining reasoning;
and curating benchmarks whose metrics expose not just final accuracy.
%

\subsection{Reward Model Design for Visual Generation}
A central obstacle for reinforcement-learning–based visual generation is the lack of a scalable and faithful reward function. 
Widely used handcrafted metrics such as FID~\cite{heusel2017gans} offer a convenient numerical signal, yet correlate only weakly with human judgments of aesthetics, semantic fidelity, or temporal coherence, especially when the task extends beyond single-frame images.
Recent learned critics, such as ImageReward~\cite{xu2023imagereward} and HPS~\cite{wu2023human} for images, and VideoReward~\cite{liu2025improving} for videos to bridge this gap by training on pairwise human-preference data, but each model targets a narrow modality and captures only a slice of perceptual quality (\eg prompt alignment or visual appeal).
As a result, policies optimized with PPO or GRPO often exploit loopholes in a single scalar signal, producing high-contrast artifacts, repetitive textures, or physically implausible motion that “game” the critic without improving real user satisfaction.
The challenge, therefore, is to design reward models that 
(i) integrate complementary low-level signals (consistency, physics, geometry) with high-level human preferences, 
(ii) generalize across images, video and 3-D scenes, and 
(iii) remain robust against reward hacking while being cheap enough to update continually as user tastes evolve.


%% file: 6_conclusion.tex
\section{Conclusion}
\label{sec:6_conclusion}

Visual reinforcement learning has transitioned from isolated proof-of-concepts to a vibrant research frontier that bridges vision, language, and action.
Our review shows that modern progress is driven by three converging forces: (i) scalable reward supervision, moving from labour-intensive RLHF to group-relative and verifiable-signal pipelines;
(ii) unified architectures, where a single policy is jointly optimised for perception, reasoning, and generation;
and (iii) ever-richer benchmarks, which measure not only task success but also alignment with human preference and policy stability.

Yet significant challenges remain. First, data and compute efficiency are pressing: current methods often require orders of magnitude more samples than supervised counterparts.
Second, robust generalization across domains, viewpoints, and embodiment settings is still limited. Third, reward design for long-horizon, open-world tasks lacks principled guidance, risking reward hacking, and unsafe behaviors.
Finally, evaluation standards must evolve to capture real-world utility, ethical alignment, and energy footprint. Addressing these issues will likely involve tighter integration of model-based planning, self-supervised visual pre-training, adaptive curricula, and safety-aware optimization.

In summary, visual RL stands poised to transform how intelligent systems perceive and interact with their surroundings. 
By unifying methodological insights and charting unresolved questions, this survey aims to serve as both a reference and a catalyst for the next wave of research toward sample-efficient, reliable, and socially aligned visual decision-making agents.